\documentclass[letterpaper]{article} 
\usepackage{aaai2026}  
\usepackage{times}  
\usepackage{helvet}  
\usepackage{courier}  
\usepackage[hyphens]{url}  
\usepackage{graphicx} 
\usepackage{natbib}  
\usepackage{caption} 
\usepackage{algorithm}
\usepackage{algorithmic}
\usepackage{tablefootnote}
\usepackage{threeparttable}

\usepackage{lipsum}
\usepackage{subcaption}
\usepackage{booktabs}
\usepackage{arydshln}
\usepackage{amsmath}
\usepackage{amsfonts}
\usepackage{multirow}
\usepackage{amssymb}
\usepackage{pifont}
\usepackage{comment}

\usepackage{newfloat}
\usepackage{listings}
\DeclareCaptionStyle{ruled}{labelfont=normalfont,labelsep=colon,strut=off} 
\floatstyle{ruled}
\newfloat{listing}{tb}{lst}{}
\floatname{listing}{Listing}
\nocopyright 

\title{InfoQ: Mixed-Precision Quantization via Global Information Flow}
\author {
    Mehmet Emre Akbulut\textsuperscript{\rm 1}
    Hazem Hesham Yousef Shalby\textsuperscript{\rm 1},
    Fabrizio Pittorino\textsuperscript{\rm 1},
    Manuel Roveri\textsuperscript{\rm 1}
}
\affiliations {
    \textsuperscript{\rm 1}Politecnico di Milano, Milan, Italy\\
    mehmetemre.akbulut@mail.polimi.it, hazemhesham.shalby@polimi.it, fabrizio.pittorino@polimi.it, manuel.roveri@polimi.it
}

\usepackage{enumitem}
\begin{document}

\maketitle

\begin{abstract}
Mixed-precision quantization (MPQ) is crucial for deploying deep neural networks on resource-constrained devices, but finding the optimal bit-width for each layer represents a complex combinatorial optimization problem. Current state-of-the-art methods rely on computationally expensive search algorithms or local sensitivity heuristic proxies like the Hessian, which fail to capture the cascading global effects of quantization error. In this work, we argue that the quantization sensitivity of a layer should not be measured by its local properties, but by its impact on the information flow throughout the entire network. We introduce InfoQ, a novel framework for mixed-precision quantization that is training-free in the bit-width search phase. InfoQ assesses layer importance by performing a single forward pass to measure the change in mutual information in the remaining part of the network, thus creating a global sensitivity score. This approach directly quantifies how quantizing one layer degrades the information characteristics of subsequent layers. The resulting scores are used to formulate bit-width allocation as an integer linear programming problem, which is solved efficiently to minimize total sensitivity under a given budget (e.g., model size or BitOps). Our retraining-free search phase provides a superior search-time/accuracy trade-off (using two orders of magnitude less data compared to state-of-the-art methods such as LIMPQ), while yielding up to a 1\% accuracy improvement for MobileNetV2 and ResNet18 on ImageNet at high compression rates (14.00$\times$ and 10.66$\times$).
\end{abstract}
\section{Introduction}
\label{sec:introduction}

The significant computational demands of Deep Neural Networks (DNNs) are a primary barrier to their deployment on resource-constrained edge devices. Mixed-precision quantization (MPQ) is a key technique to overcome this, assigning low bit-widths to robust layers while preserving precision for sensitive ones to outperform uniform quantization \cite{Dong2023EMQET}. However, identifying the optimal bit-width configuration for a network is a combinatorial problem; the search space grows exponentially with network depth, making an exhaustive search infeasible.

Current approaches to this challenge fall into two categories, both of which suffer from fundamental limitations. \textit{Search-based} methods \cite{wang2019haqhardwareawareautomatedquantization, wu2018mixedprecisionquantizationconvnets} use expensive reinforcement learning or neural architecture search (NAS) to find good configurations. While sometimes effective, they are computationally prohibitive and treat the network as a black box, failing to build a principled model of quantization impact.
More efficiently, \textit{criterion-based} methods use a proxy metric to score layer sensitivity in a single shot. A relevant class of these methods use the trace of the Hessian \cite{dong2019hawqhessianawarequantization, dong2019hawqv2hessianawaretraceweighted}. The intuition is that layers in \textit{flat} regions of the loss landscape are less sensitive. However, the prohibitive cost of the full Hessian forces these methods to use a \textit{block-diagonal approximation}, which explicitly ignores all inter-layer dependencies. This makes the Hessian-based analysis fundamentally local, failing to capture how quantization error in one layer cascades and degrades the representational quality of subsequent layers. Other criteria, like learned step-sizes \cite{tang2023mixedprecisionneuralnetworkquantization}, are similarly layer-local and often require expensive retraining. This leaves a critical gap: a fast, principled MPQ method that evaluates sensitivity based on its global impact across layers.

In this work, we argue that the true sensitivity of a layer is not an intrinsic, local property, but is defined by the \textit{global impact} it has on the \textit{information flow throughout the entire network}. We introduce \textit{InfoQ}, a framework that directly builds upon this principle. Instead of using local proxies, we ask a more fundamental question: \textit{How does quantizing a layer degrade the ability of downstream layers to represent critical information about the input and the final task?}

To answer this, \textit{InfoQ} leverages tools from information theory. For each layer and candidate bit-width, we perform a single forward pass and measure the resulting change in mutual information (MI) at subsequent \textit{observer} layers. This change directly quantifies the global, cascading effect of the quantization error. These global sensitivity scores are then used to formulate the bit-width allocation as an Integer Linear Programming (ILP) problem, which is solved efficiently to find the optimal configuration under a given resource budget (e.g., model size or BitOps). Our search phase is entirely training-free and provides a global, interpretable sensitivity measure, in contrast to existing local methods.
\begin{table*}[t!]
\small
    \centering
    \begin{tabular}{c c c c c c c}
    \toprule
    \textbf{Method} & \textbf{AutoQ} & \textbf{DNAS} & \textbf{LIMPQ} & \textbf{HAWQv2} & \textbf{MPQCO} & \textbf{\textit{InfoQ} (Ours)} \\
    \midrule
    Iterative Search Avoidance & \ding{55} & \ding{55} & \ding{51} & \ding{51} & \ding{51} & \ding{51} \\
    Unlimited Search Space & \ding{51} & \ding{55} & \ding{51} & \ding{51} & \ding{55} & \ding{51} \\
    Training-Free Search & \ding{55} & \ding{55} & \ding{55} & \ding{51}& \ding{51} & \ding{51} \\
    Fully Automatic Bit Assignment & \ding{51} & \ding{51} & \ding{51} & \ding{51} & \ding{55} & \ding{51} \\
    Global Impact Metric & \ding{55} & \ding{55} & \ding{55} & \ding{55} & \ding{55} & \ding{51} \\
    \bottomrule
    \end{tabular}
    \caption{Comparison of key properties in state-of-the-art MPQ methods. \textit{InfoQ} is the only method that combines a training-free search with a global sensitivity metric, enabling fully automatic bit allocation that accounts for network-wide error propagation.
    }
    \label{tab:comparison}
\end{table*}
Our contributions are:
\begin{enumerate}
    \item We introduce \textit{InfoQ}, the first MPQ framework to use a global sensitivity metric based on information flow, directly capturing the cascading effects of quantization error ignored by local methods.
    \item Our method achieves state-of-the-art accuracy, recovering full-precision accuracy on ResNet18 and narrowing the gap to just 0.34\% on ResNet50 at high compression rates, all while using orders of magnitude less data for the search phase than competing methods at the same accuracy level.
    \item Using a practical method for mutual information estimation, we analyze deep networks compression from an information-theoretic perspective, highlighting a valuable application for both the model compression and information theory literature.
\end{enumerate}

\section{Related Work}
\label{sec:rw}
Mixed-precision quantization (MPQ) consistently demonstrates superior performance over uniform-precision approaches \cite{Dong2023EMQET}. The central challenge in MPQ is navigating the combinatorial search space of bit-width configurations. Methodologies to solve this problem are broadly categorized as \textit{search-based} and \textit{criterion-based}.

The \textit{search-based} methods 
solve MPQ problem with computationally intensive search algorithms. Reinforcement learning (RL) has been employed to select layer-wise \cite{wang2019haqhardwareawareautomatedquantization} or even kernel-wise \cite{lou2020autoqautomatedkernelwiseneural} bit allocations. Another explored direction adapts techniques from Neural Architecture Search (NAS), such as differentiable search over a super-network of quantization operators \cite{wu2018mixedprecisionquantizationconvnets, Cai2020RethinkingDS, yu2020searchwantbarrierpanelty}. While these methods can yield high-performing configurations, their reliance on costly, iterative search (often requiring a significant number of GPU-hours) and large validation datasets makes them impractical for many applications. Our work bypasses this expensive search phase entirely.

To circumvent the cost of search, \textit{criterion-based} methods use an efficient proxy metric for layer sensitivity to guide a one-shot bit allocation. Methods like HAWQ \cite{dong2019hawqhessianawarequantization} and HAWQv2 \cite{dong2019hawqv2hessianawaretraceweighted} use the trace or eigenvalues of the Hessian as a proxy for layer sensitivity, using second-order information from the loss landscape. As noted in the Introduction, these methods are constrained to a block-diagonal approximation of the Hessian, making their analysis fundamentally local. Other works have proposed alternative proxies, including layer-wise orthogonality metrics \cite{Ma2021OMPQOM} and the learned scale parameters of quantizers \cite{tang2023mixedprecisionneuralnetworkquantization}. While significantly faster than search-based methods, these approaches share a critical limitation: their reliance on layer-local heuristics prevents them from quantifying the global, cascading effects of quantization error. \textit{InfoQ} is a criterion-based method, but its criterion is, by design, global.

Concerning the grounding of model compression in information theory, it is a promising direction for developing more principled methods. The Information Bottleneck (IB) principle, in particular, offers a theoretical framework for analyzing the trade-off between compression and predictive accuracy \cite{Tishby2015DeepLA}. The IB framework has been successfully applied to guide network pruning, where channels or weights are removed based on their information content \cite{Guo2023AutomaticNP, Zheng2021AnIT, Nielsen2021CompressionOD}. However, its application to MPQ has remained limited. 

Our work, \textit{InfoQ}, is a criterion-based method that, for the first time, successfully uses information-theoretic tools to construct a \textit{practical global sensitivity criterion} for MPQ. Unlike prior IB-based model pruning methods, we do not attempt a complex optimization of the IB objective. Instead, we use mutual information as a direct, interpretable metric to measure the end-to-end impact of quantization. As summarized in Table~\ref{tab:comparison}, \textit{InfoQ} is the first MPQ framework able to provide a fully automatic bit allocation based on a training-free, interpretable, and global analysis of the network information flow.

\section{Background: Estimating Mutual Information in Deep Networks}
\label{sec:background}

To analyze the global impact of quantization, we turn to concepts from information theory, particularly those used by the Information Bottleneck (IB) principle~\cite{Tishby2015DeepLA}. The IB principle models a DNN as a system that learns compressed representations of an input $X$. The quality of a representation $L_i$ at layer $i$ is evaluated by two fundamental quantities: the mutual information it retains about the input, $I(X;L_i)$, and the information it preserves about the final target, $I(L_i;Y)$ \cite{Geiger2020OnIP}. 
While the IB principle aims to optimize the trade-off between these two, we leverage them as direct probes to measure the information variations caused by quantization along the network.
For two continuous random variables $U$ and $V$, the mutual information $I(U;V)$ quantifies their statistical dependency and is defined as $I(U;V) = \int_{\mathcal{V}} \int_{\mathcal{U}} p(u,v) \log \frac{p(u,v)}{p(u)p(v)} du dv$, where $p(u,v)$ is the joint probability density function and $p(u)$ and $p(v)$ are the marginal densities. The quantities $I(X;L_i)$ and $I(L_i;Y)$ are the core metrics for quantifying the informational properties of a network representations.

There are however fundamental challenges in estimating mutual information for high-dimensional, deterministic functions like DNNs. The first obstacle is that the activations $L_i$ are a deterministic function of the input $X$. In a continuous setting, this leads to a theoretically infinite mutual information, $I(X;L_i)$, making the metric uninformative \cite{Amjad2018LearningRF}.
The second is that, even when non-determinism is introduced (e.g., through quantization \cite{Lorenzen2021InformationBE} or noise injection), the high dimensionality of both the input space $\mathcal{X}$ and the activation space $\mathcal{L}_i$ makes non-parametric MI estimation intractable (the \textit{curse of dimensionality}): the number of samples required for a reliable estimate grows exponentially with data dimension~\cite{Goldfeld2019ConvergenceOS}.

To overcome these challenges, our work leverages two key techniques from recent literature on tractable MI estimation. The first is \textit{lossy compression} of the input space \cite{Butakov2023InformationBA}, where a high-dimensional input~$X$ (e.g., an image) is mapped to a lower-dimensional, semantically rich feature vector using a powerful pre-trained encoder. This reduces the dimensionality of one of the variables, mitigating the estimation problem.
The second central technique is \textit{Sliced Mutual Information} (SMI), a computationally efficient and scalable surrogate for MI~\cite{Goldfeld2022kSlicedMI}. Instead of estimating the MI between two high-dimensional random vectors directly, SMI computes the average MI between one-dimensional random projections of these vectors. For vectors $U \in \mathbb{R}^d$ and $V \in \mathbb{R}^p$, SMI is defined as
$
    \text{SMI}(U;V) = \mathbb{E}_{\theta_u, \theta_v} \left[ \text{MI}(\theta_u^T U; \theta_v^T V)\right]
$,
where the projection directions $\theta_u$ and $\theta_v$ are drawn randomly from unit spheres. The MI between these one-dimensional scalar projections can be robustly and efficiently estimated using standard non-parametric methods, such as 
Kraskov–Stögbauer–Grassberger (KSG) estimator \cite{Kraskov2003EstimatingMI}.

\section{Methodology}

\subsection{Problem Formalization}
Let a deep neural network be defined by a sequence of $L$ layers with full-precision parameters $\boldsymbol{W}$. For mixed-precision quantization, we define a discrete set of candidate bit-widths $\boldsymbol{B} = \{b_0, b_1, \ldots, b_{n-1}\}$. For each layer $\ell \in \{1, \dots, L\}$, we assign a bit-width for its weights, $b_w^{(\ell)} \in \boldsymbol{B}$, and its activations, $b_a^{(\ell)} \in \boldsymbol{B}$.
A complete bit-width configuration for the network is an assignment vector $\boldsymbol{s} = \{ (b_w^{(1)}, b_a^{(1)}), \dots, (b_w^{(L)}, b_a^{(L)}) \}$. The set of all possible configurations, $\mathcal{A}$, forms the search space for the MPQ problem. For a typical network with $L=50$ and $|\boldsymbol{B}|=4$ candidate bit-widths for both weights and activations, the size of this search space is $|\mathcal{A}| = (|\boldsymbol{B}|^2)^L = 16^{50}$.

The objective of MPQ is to find an optimal bit-width configuration $\boldsymbol{s}^* \in \mathcal{A}$ that minimizes a task-specific loss $\mathcal{L}$ subject to one or more resource constraints:
\begin{equation}
    \label{eq:mpq_problem}
    \begin{aligned}
    \boldsymbol{s}^* = \underset{\boldsymbol{s} \in \mathcal{A}}{\arg\min} & \quad \mathcal{L}(f(\boldsymbol{x}; \boldsymbol{s}, \boldsymbol{W_{\boldsymbol{s}}}), \boldsymbol{y}) \\
    \textrm{subject to} & \quad \text{Cost}(\boldsymbol{s}) \leq C
    \end{aligned}
\end{equation}
where $f(\cdot)$ is the network model with parameters $\boldsymbol{W_{\boldsymbol{s}}}$ and activations quantized according to configuration $\boldsymbol{s}$, $(\boldsymbol{x}, \boldsymbol{y})$ are the data and labels, and $\text{Cost}(\boldsymbol{s})$ represents a resource budget. Common cost functions include model size \cite{Uhlich2019MixedPD} or computational complexity, measured in BitOps \cite{Yang2020FracBitsMP}:
\begin{equation}
    \label{eq:cost_functions}
    \text{Cost}(\boldsymbol{s}) = \sum_{\ell=1}^{L} \text{Size}(b_w^{(\ell)}) \quad \text{or} \quad \sum_{\ell=1}^{L} \text{BitOps}(b_w^{(\ell)}, b_a^{(\ell)})
\end{equation}

\subsection{The \textit{InfoQ} Method}
Directly solving the optimization problem in Eq.~\ref{eq:mpq_problem} is intractable: evaluating the loss for each candidate $\boldsymbol{s}$ would require retraining or fine-tuning the model, and the combinatorial size of $\mathcal{A}$ makes exhaustive search impossible. 
We therefore propose a proxy-based method to efficiently estimate the impact of a given bit-width configuration without full model retraining, \textit{InfoQ}, which reformulates the problem as a three-step process.
First, it defines a layer quantization sensitivity as the global information degradation it causes throughout the network. Then, it computes this sensitivity for each layer and candidate bit-width using an efficient SMI-based algorithm. Finally, it uses these scores to formulate the bit-width allocation as an Integer Linear Programming (ILP) problem, which can be solved efficiently for any given resource constraint.

\subsubsection{Defining Sensitivity via Global Information Flow}
Our central hypothesis is that the impact of quantizing a layer is not a local phenomenon but is best measured by its effect on the information propagated through subsequent layers. We quantify this impact by measuring the change in the two fundamental information-theoretic quantities analyzed in the IB context:
the \textit{input information} $I(X; L_j)$, which measures how much information the representation $L_j$ at a downstream layer $j$ retains about the original input $X$; and the \textit{task-relevant information} $I(L_j; Y)$, which measures how much information $L_j$ contains about the final task labels $Y$.
A significant change in either of these quantities at a downstream layer $j$ after quantizing an upstream layer $i$ indicates that layer $i$ is sensitive to quantization.
To make the estimation of these quantities tractable for high-dimensional data, we employ the SMI surrogate, as introduced in Section~\ref{sec:background}. 
Furthermore, we apply \textit{lossy compression} where input images $X$ are mapped to a lower-dimensional embedding $X_{\mathcal{E}}$ using a pre-trained DINOv2 encoder~\cite{Oquab2023DINOv2LR} that we denote by~$\mathcal{E}$, i.e. $X_{\mathcal{E}} = \mathcal{E}(X)$. 

Our sensitivity metric is therefore based on the measured informational degradation in $\text{SMI}(X_{\mathcal{E}}; L_j)$ and $\text{SMI}(L_j; Y)$ relative to a high-precision baseline model (all layers at 8-bit).
Let $L_{j, \text{8bit}}$ be the activations of a downstream \textit{observer} layer~$j$ in the baseline model, and let $L_{j, b}$ be the activations at the same layer when an upstream layer $i$ has been quantized to bit-width $b$. We formally define the absolute change in SMI as:
\begin{equation}    
\begin{aligned}
    \Delta \text{SMI}_{X,L}^{(i,b,j)} &= |\text{SMI}(X_{\mathcal{E}}; L_{j, \text{8bit}}) - \text{SMI}(X_{\mathcal{E}}; L_{j, b})|\\
    \Delta \text{SMI}_{L,Y}^{(i,b,j)} &= |\text{SMI}(L_{j, \text{8bit}}; Y) - \text{SMI}(L_{j, b}; Y)|
\end{aligned}
\label{eq:deltaSMI}
\end{equation}
The total sensitivity score for quantizing layer $i$ to bit-width~$b$ is the normalized sum of these measured changes across a set of pre-selected \textit{observer layers}.

\begin{figure}[t]
    \centering
    \includegraphics[width=0.495\columnwidth]{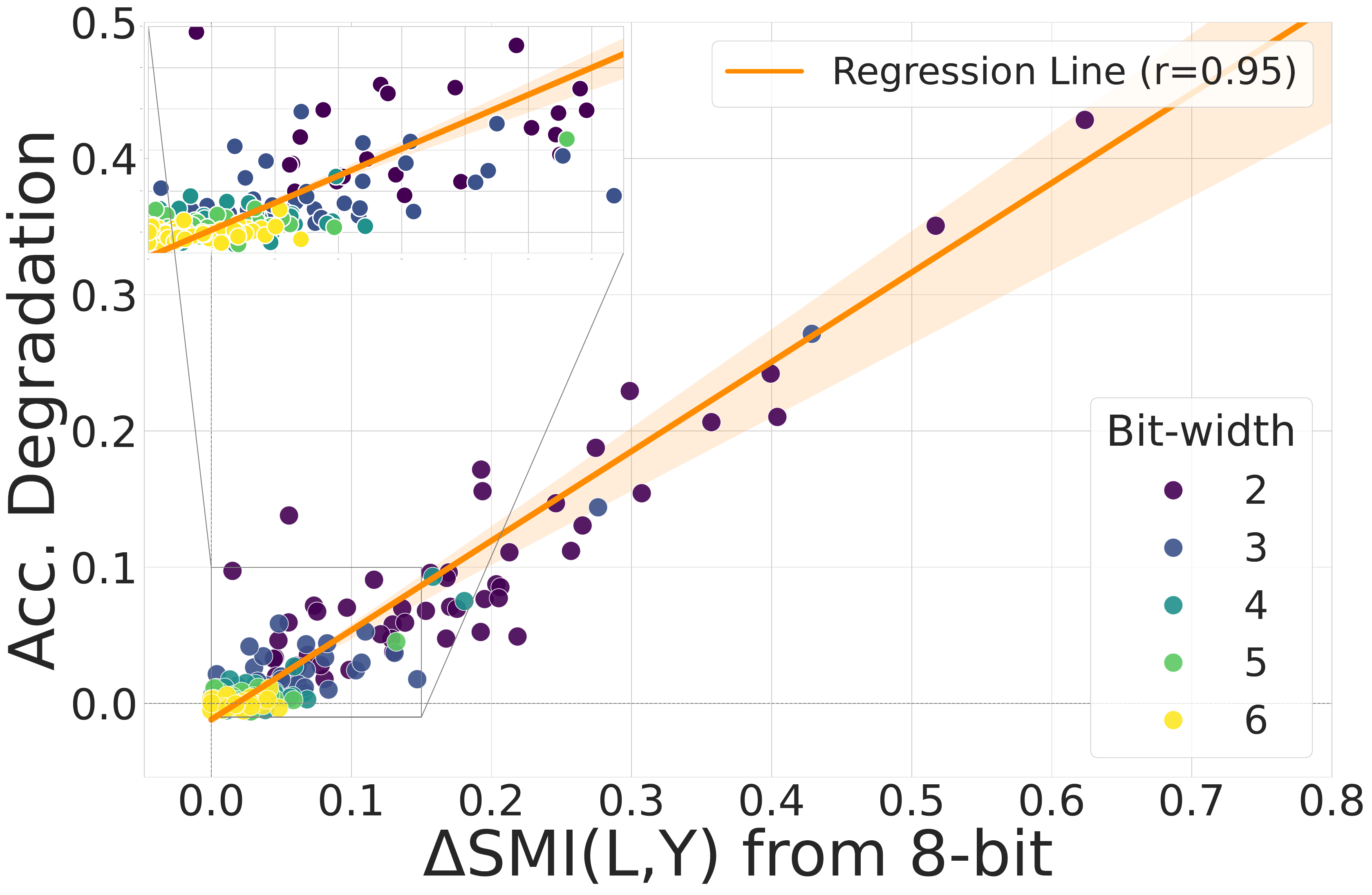}    \includegraphics[width=0.495\columnwidth]{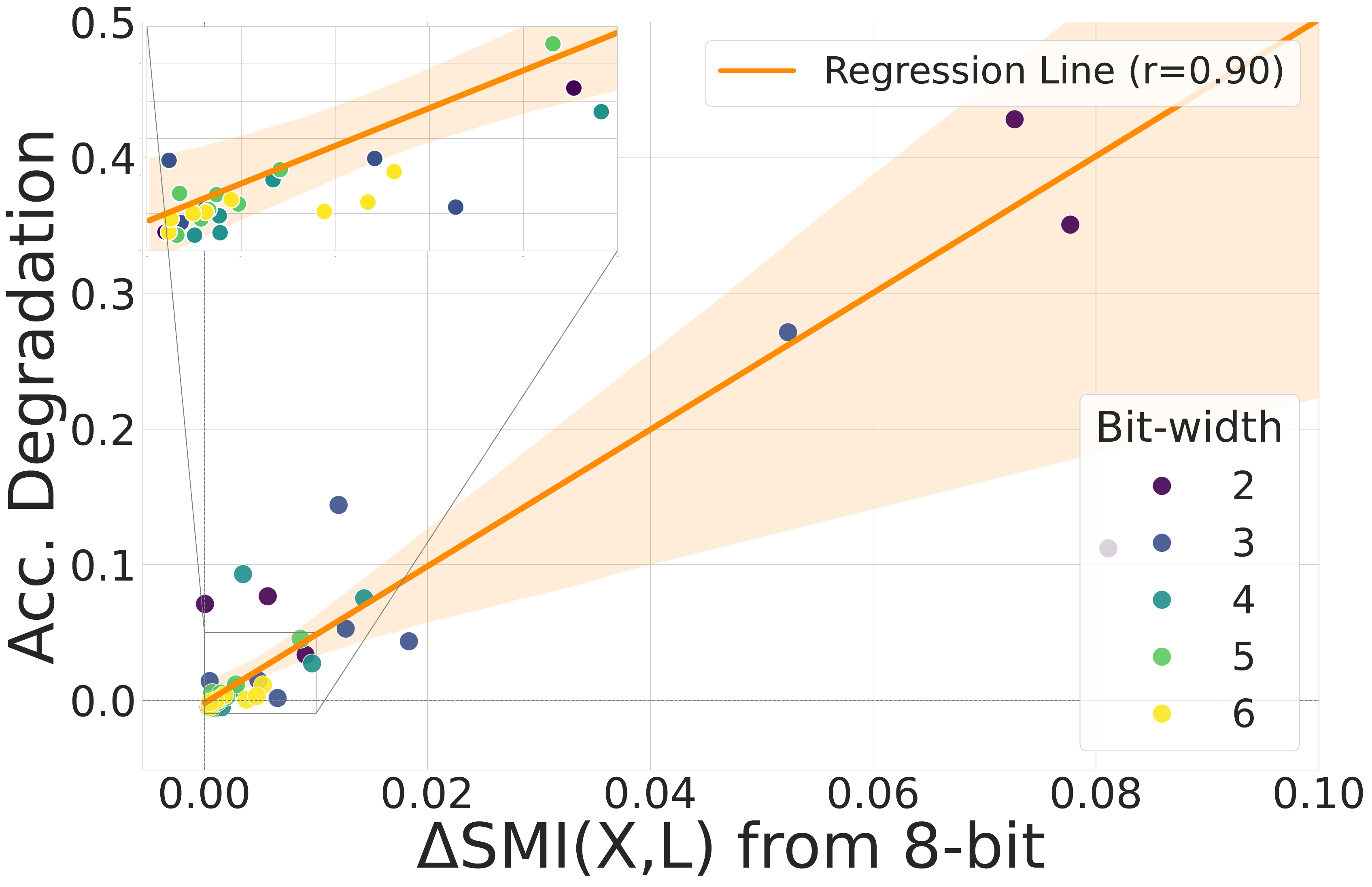}
    \caption{Correlation analysis for $\Delta$SMI metrics defined in Eqs.~\ref{eq:deltaSMI} and accuracy degradation on ResNet50 on ImageNet. The accuracy drop is shown in function of (left panel) the target-relevant information change, $\Delta \text{SMI}(L,Y)$, on the \textit{global average pooling} layer, and (right panel) the input-relevant information change, $\Delta \text{SMI}(X,L)$, on the \textit{conv} layer at index 7. The strong positive correlation demonstrates that the $\Delta$SMI metrics are an effective proxy for quantization sensitivity.
    }
    \label{fig:smi_corr}
\end{figure}
\subsubsection{Observer Layer Selection and Validation}

A critical component of our method is the selection of downstream \textit{observer layers}, i.e. the locations where we measure the the absolute change in SMI in Eqs.~\ref{eq:deltaSMI}. An ideal observer layer is one where a change in its informational content is highly predictive of the final degradation in task performance. We identify these observers for a given network architecture via a one-time, empirical correlation analysis.
For each layer $i$ in the network, we quantize it to a low bit-width ($<$8-bit) while keeping all other layers at 8-bit. We then perform a forward pass and record two values: (1) the final top-1 accuracy degradation, and (2) $\Delta \text{SMI}_{X,L}$ and $\Delta \text{SMI}_{L,Y}$ as per Eqs.~\ref{eq:deltaSMI}, at all subsequent layers $j > i$. By repeating this for all quantizable layers, 
we are able to compute the Pearson correlation coefficient between the absolute change in SMI as per Eqs.~\ref{eq:deltaSMI} and the final accuracy drop. This allows us to identify a robust set of observers by selecting layers strongly correlated with the final accuracy drop (Pearson’s $r > 0.70$).
This analysis, detailed in the Appendix, yields a clear and consistent heuristic. We find that the change in task-relevant information, $\Delta \text{SMI}_{L,Y}$, measured at layers toward the end of the network (e.g., global pooling and fully-connected layers) exhibits the strongest correlation with accuracy degradation. Conversely, the change in input-relevant information, $\Delta \text{SMI}_{X,L}$, is most predictive when measured at intermediate layers. 

\begin{figure}[t]
    \centering
    \begin{subfigure}{0.48\textwidth}
        \centering
        \begin{subfigure}{0.49\textwidth}
                \includegraphics[width=1\linewidth]{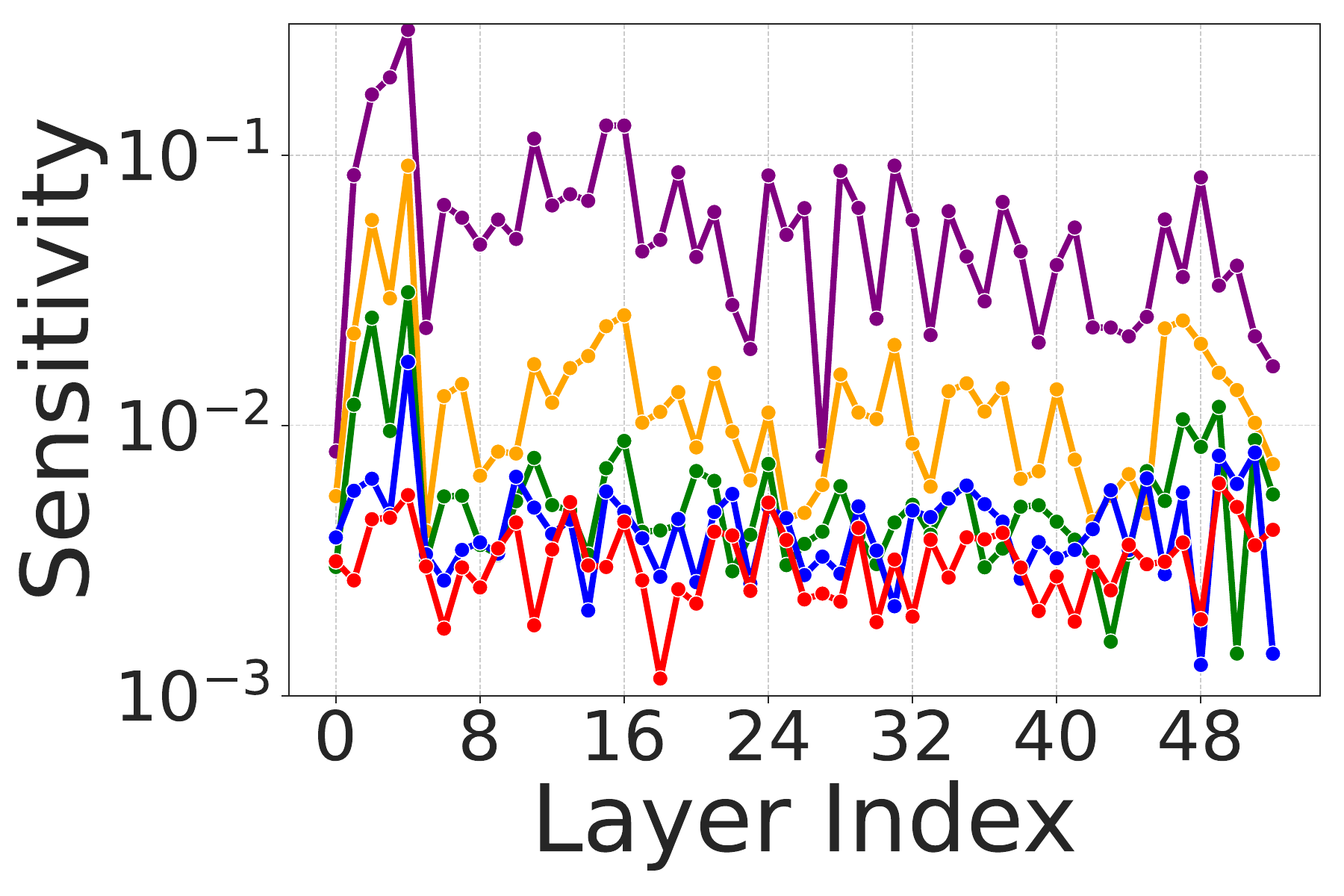}\hfill
        \caption{ResNet50 Weights}
        \end{subfigure}
        \begin{subfigure}{0.49\textwidth}
            \includegraphics[width=1\linewidth]{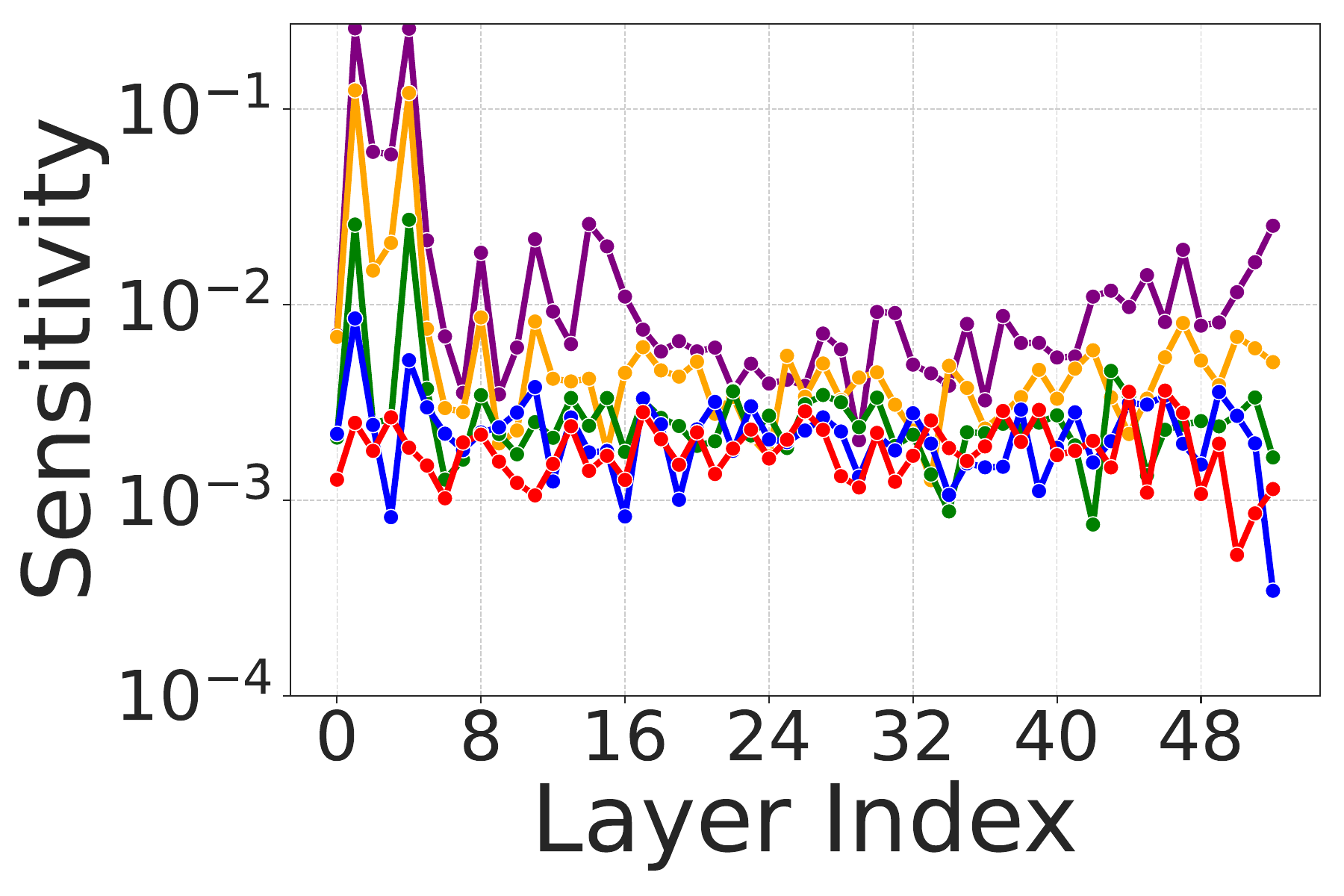}
        \caption{ResNet50 Activations}
        \end{subfigure}
    
        \label{fig:resnet50_group} 
    \end{subfigure}
    \begin{subfigure}{0.48\textwidth}
        \centering
        \begin{subfigure}{0.49\textwidth}
                \includegraphics[width=1\linewidth]{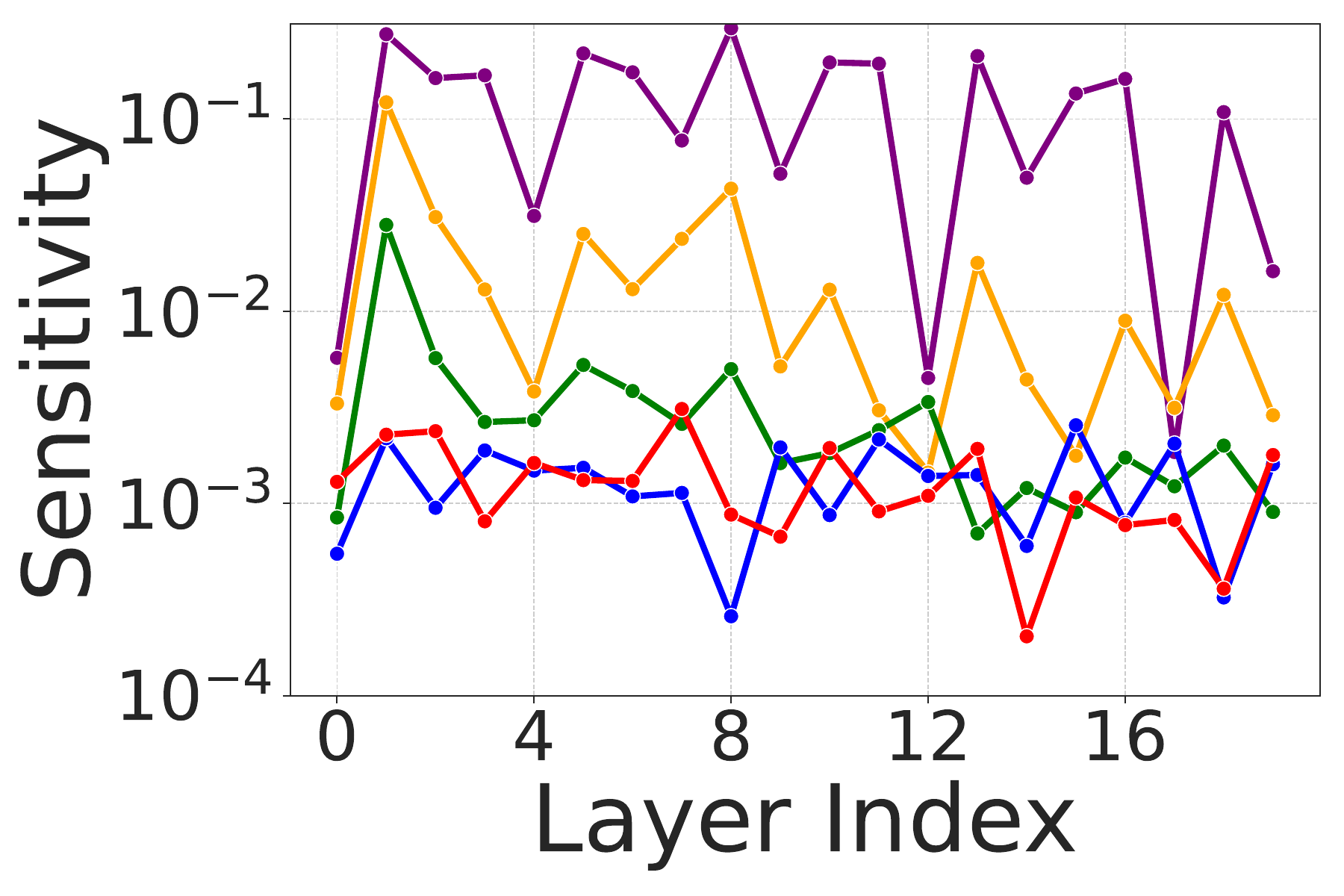}\hfill
        \caption{ResNet18 Weights}
        \end{subfigure}
        \begin{subfigure}{0.49\textwidth}
            \includegraphics[width=1\linewidth]{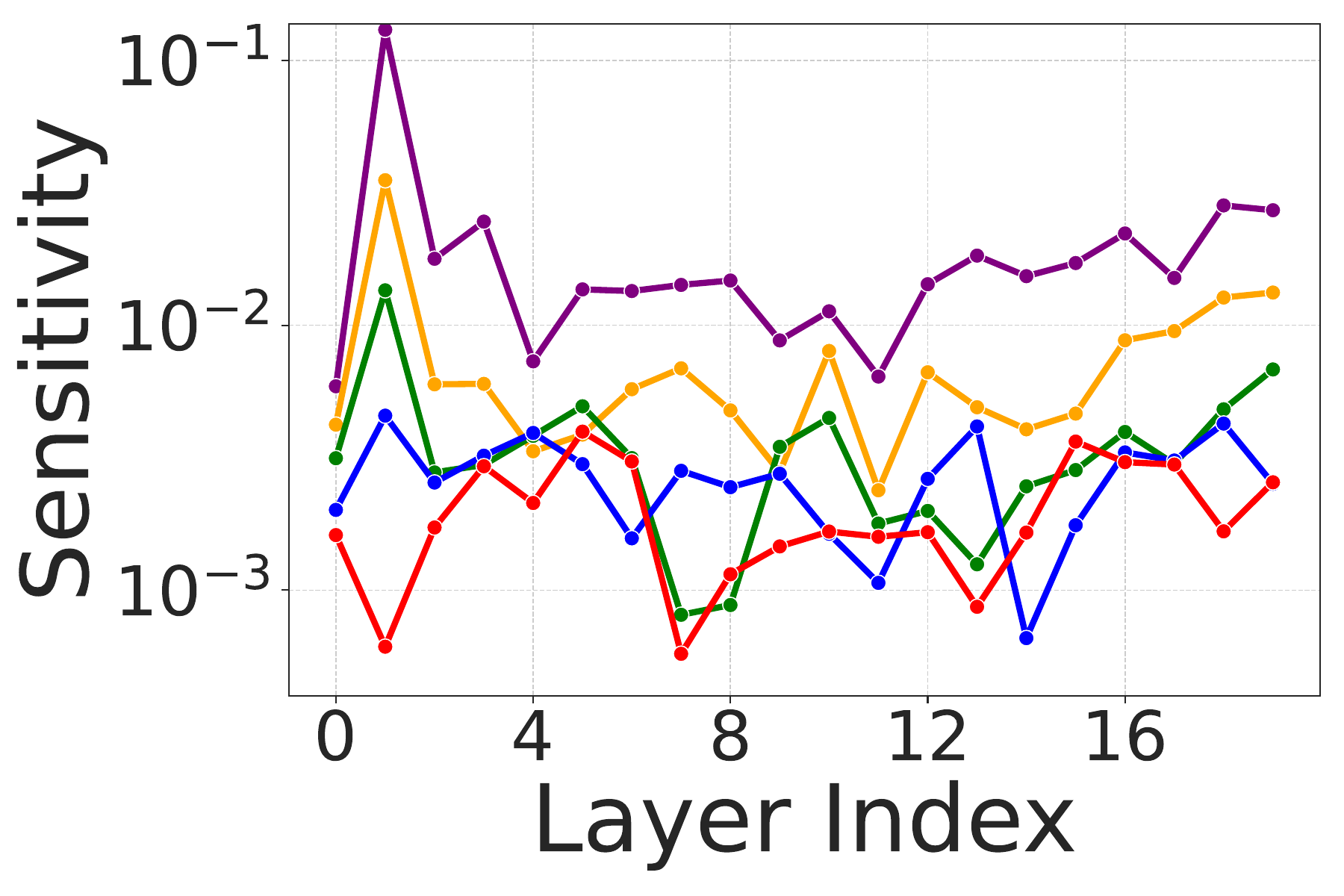}
        \caption{ResNet18 Activations}
        \end{subfigure}
    \begin{subfigure}{0.72\textwidth}
    \centering
    \includegraphics[width=1\linewidth]{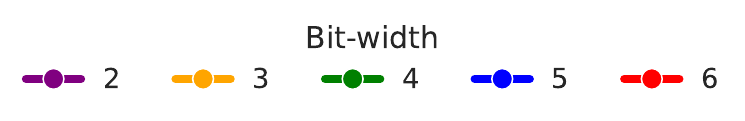}
    \end{subfigure}
        \label{fig:resnet18_group} 
    \end{subfigure}
    \hfill 
    \caption{Sensitivity profiles as per Eq.~\ref{eq:score} for ResNet50 and ResNet18 on ImageNet in function of the layer index, for different quantization bit-widths.
    }
    \label{fig:sensitivities}
\end{figure}

As shown in Fig.~\ref{fig:smi_corr}, the absolute change in SMI measured at selected observer layers strongly correlates with the final model accuracy, validating their use as an effective proxy for quantization sensitivity. Fig.~\ref{fig:sensitivities} visualizes the final sensitivity scores computed for each layer across various bit-widths, which form the basis of our bit allocation strategy.

\subsubsection{Sensitivity Scoring and Optimal Bit Allocation}

With the observer layers identified, we proceed to compute a sensitivity score for each quantizable layer and candidate bit-width. This process, detailed in Algorithm~\ref{alg:smi_computation}, is designed to be efficient and training-free.
The core of the algorithm is a systematic measurement of informational degradation. We first establish a baseline by computing the values $\text{SMI}(X_{\mathcal{E}}; L_j)$ and $\text{SMI}(L_j; Y)$ at all identified observer layers~$j$ and~$k$ (notice that in general they differ) for a reference model where all layers are quantized to 8-bits. Then, to compute the sensitivity of a target layer $i$ for a candidate bit-width $b \in \boldsymbol{B}$, we perturb this baseline: we quantize only layer~$i$ to bit-width~$b$ and perform a single forward pass over a small calibration dataset. The sensitivity score is then the normalized sum of the absolute change
in SMI (as defined in Eqs.~\ref{eq:deltaSMI}) across all downstream observer layers. This procedure is repeated for every (layer, bit-width) pair.
Therefore, the final score $S$ for quantizing the parameters of layer $i$ (weights or activations) to bit-width $b$ is formulated as:
\begin{align}
    \label{eq:score}
     &S(i, b, \mathcal{O}_{XL}^{j>i}, \mathcal{O}_{LY}^{j>i}) = \frac{1}{b} \frac{\langle\Delta\text{SMI}\rangle_{\mathcal{O}^{j>i}}}{\langle\text{SMI}\rangle^{\text{8bit}}_{\mathcal{O}^{j>i}}} \\
    &\langle\Delta\text{SMI}\rangle_{\mathcal{O}^{j>i}} =  \displaystyle\sum_{\substack{j\in \mathcal{O}_{XL}^{j>i}}}  \Delta \text{SMI}_{X,L}^{(i,b,j)} + \displaystyle\sum_{\substack{k\in \mathcal{O}_{LY}^{j>i}}} \Delta \text{SMI}_{L,Y}^{(i,b,k)} \nonumber \\
    &\langle\text{SMI}\rangle^{\text{8bit}}_{\mathcal{O}^{j>i}} =  \sum_{j\in \mathcal{O}_{XL}^{j>i}} \text{SMI}_{X,L}^{(j), \text{8bit}} +
     \sum_{k\in \mathcal{O}_{LY}^{j>i}}\text{SMI}_{L,Y}^{(k), \text{8bit}} \nonumber
\end{align}
where $\mathcal{O}_{XL}^{j>i}$ and $\mathcal{O}_{LY}^{j>i}$ are the downstream observer layers (with respect to layer $i$) for $\text{SMI}(X_{\mathcal{E}}; L_j)$ and $\text{SMI}(L_j; Y)$, respectively. The score is normalized by the total baseline information to ensure comparability across different observer sets. We also include a $1/b$ term to penalize lower bit-widths, in order to explicitly account for their inherently higher sensitivity to quantization error. When the ILP solver seeks to minimize total network sensitivity under a budget, this penalty ensures that, for a comparable $\Delta$SMI value, a higher bit-width configuration is favored, promoting a more stable quantization setup. We study the impact of the factor $1/b$ in the Appendix, showing that  the global SMI metric is the dominant factor in our method performance.

Once the sensitivity scores $S_w(\ell, b)$ and $S_a(\ell, b)$ are computed for all layers and bit-widths, we can find the optimal network bit-width configuration $\boldsymbol{s}^*$ by solving a constrained optimization problem. The goal is to find the configuration that minimizes the total network sensitivity while respecting a resource budget $C$. This is formulated as a classic Integer Linear Programming (ILP) problem:
\begin{equation}
\label{eq:ilp}
\begin{aligned}
    \boldsymbol{s}^* = \underset{\boldsymbol{s} \in \mathcal{A}}{\arg\min} \quad & \sum_{\ell=1}^{L} \left( S_w(\ell, b_w^{(\ell)}) + \alpha S_a(\ell, b_a^{(\ell)}) \right) \\
    \text{subject to} \quad & \text{Cost}(\boldsymbol{s}) \leq C
\end{aligned}
\end{equation}
Here, $\alpha$ is a hyperparameter balancing the relative importance of weight versus activation quantization sensitivity. $\text{Cost}(\boldsymbol{s})$ is the resource-specific cost function (e.g., model size or BitOps, as in Eq.~\ref{eq:cost_functions}). 
For only two experiments on ResNet18 and ResNet50, we
set $\alpha = 0.1$ and $\alpha = 2$, respectively, to ensure a fair comparison with our main competitor, LIMPQ \cite{tang2023mixedprecisionneuralnetworkquantization}. Other experiments on ResNet18 and MobileNetV2 employ weight-only MPQ which does not need hyperparameter $\alpha$.
This standard ILP problem is solved efficiently using an off-the-shelf solver \cite{Mitchell2011PuLPA}, making the final allocation step extremely fast for any given budget~$C$.

\begin{table*}[t]
\small
    \centering
    \begin{threeparttable}
    \begin{tabular}{ccccccccc} 
        \toprule
        Method & W-bit & A-bit & Top-1/Full & Top-1/Quant & Top-1/Drop & BitOps(G) & Searching Data\\
        \midrule
        LQ-Nets \cite{zhang2018lqnetslearnedquantizationhighly} & 3 & 32 & 70.30 & 69.30 & -1.00 & - & -  \\
        LQ-Nets \cite{zhang2018lqnetslearnedquantizationhighly} & 4 & 32 & 70.30 & 70.00 & -0.30 & - & - \\
        OMPQ \cite{Ma2021OMPQOM} & 3MP & 8 & 71.08 & 69.94 & -1.14 & - & \textbf{64} \\
        MPQCO \cite{chen2021mixedprecisionquantizationneuralnetworks} & 3MP & 32 & 69.76 & 69.50 & -0.26 & - & 1024 \\
        MPQCO \cite{chen2021mixedprecisionquantizationneuralnetworks} & 3MP & 8 & 69.76 & 69.39 & -0.37 & - & 1024\\
        \cdashline{1-8} 
        \textbf{InfoQ (Ours)} & \textbf{3MP} & \textbf{8} & \textbf{70.60} & \textbf{70.94} & \textbf{+0.34} & & 4800 \\
        \midrule 
         PACT \cite{choi2018pactparameterizedclippingactivation} & 3 & 3 & 70.40 & 68.10 & -2.30 & 23.09 & -  \\
         LQ-Nets \cite{zhang2018lqnetslearnedquantizationhighly} & 3 & 3 & 70.30 & 68.20 & -2.10 & 23.09 & - \\
         Nice \cite{Baskin_2021} & 3 & 3 & 69.80 & 67.70  & -2.10 & 23.09 & - \\
         AutoQ \cite{lou2020autoqautomatedkernelwiseneural} & 3MP & 3MP & 69.90 & 67.50 & -2.40 & - & - \\
         SPOS \cite{guo2020singlepathoneshotneural} & 3MP & 3MP & 70.90 & 69.40 & -1.50 & 21.92 & - \\
         DNAS \cite{wu2018mixedprecisionquantizationconvnets} & 3MP & 3MP & 71.00 & 68.70 & -2.30 & 25.38 & -  \\
         LIMPQ\tnote{1} \cite{tang2023mixedprecisionneuralnetworkquantization} & 3MP & 3MP & 70.50 & 69.70 & - 0.80 & 23.07 & 0.6M \\
         \cdashline{1-8}
         \textbf{InfoQ (Ours)} & \textbf{3MP} & \textbf{3MP} & \textbf{70.60} & \textbf{69.99} & \textbf{-0.61} & \textbf{23.04} & \textbf{4800} \\
        \bottomrule
    \end{tabular}
        \begin{tablenotes}
            \item[1] First layer activation is not quantized. 
        \end{tablenotes}
    \end{threeparttable}
     \caption{Results for ResNet18 on ImageNet under BitOps constraints. Upper part is weight-only, below part is non-limited MPQ. `W-bit' and `A-bit' represent the bit-widths of weights and activations, respectively. `MP' denotes mixed-precision quantization. `Top-1/Quant' and `Top-1/Full' refer to the top-1 accuracy of the quantized and full-precision models. `Top-1/Drop' is defined as the accuracy drop: Top-1/Full - Top-1/Quant.
     }
\label{tab:resnet_18_imagenet}
\end{table*}
\begin{table*}[t!]
    \small
    \begin{threeparttable}
    \begin{tabular}{cccccccccc} 
        \toprule
        Method & W-bit & A-bit & Top-1/Full & Top-1/Quant & Top-1/Drop &  W-C & Searching Data\\
        \midrule
         PACT \cite{choi2018pactparameterizedclippingactivation} & 3 & 3 & 76.90 & 75.30 & -1.60 & 10.67$\times$ & - \\
         LQ-Nets \cite{zhang2018lqnetslearnedquantizationhighly} & 3 & 3 & 76.00 & 74.20 & -1.80 & 10.67$\times$& - \\
         HAQ \cite{wang2019haqhardwareawareautomatedquantization} & 3MP & 8 & 75.30 & 76.20 & -0.90 & 10.57$\times$ & -  \\
         DiffQ \cite{défossez2022differentiablemodelcompressionpseudo} & 3MP & 32 & 77.10 & 76.30 & -0.80 & 11.10$\times$& - \\
         BP-NAS \cite{yu2020searchwantbarrierpanelty} & 4MP & 4MP & 77.50 & 76.70 & -0.80 & \textbf{11.10}$\times$& - \\
         HAWQ \cite{dong2019hawqhessianawarequantization} & MP & MP & 77.30 & 75.50 & -1.80 & 12.20$\times$& - \\
         HAWQv2 \cite{dong2019hawqv2hessianawaretraceweighted} & MP & MP & 77.30 & 75.80 & -1.50 & 12.20$\times$& \textbf{256}  \\
         MPQCO \cite{chen2021mixedprecisionquantizationneuralnetworks} & MP & 4MP & 76.10 & 75.30 & -0.80 & 12.20$\times$ & 1024  \\
         LIMPQ\tnote{1} \cite{tang2023mixedprecisionneuralnetworkquantization} & 3MP & 4MP & 77.50 & 76.90 & -0.60 & 12.20$\times$& 0.6M  \\
        \cdashline{1-10}    
        \textbf{InfoQ (Ours)} & \textbf{3MP} & \textbf{4MP} & \textbf{77.37} & \textbf{77.03} & \textbf{-0.34} & 12.20$\times$ & 6000 \\
        \bottomrule
    \end{tabular}
            \begin{tablenotes}
            \item[1] First layer activation is not quantized. 
        \end{tablenotes}
    \end{threeparttable}
     \centering
    \caption{Results for ResNet50 on ImageNet under BitOps constraints. 'W-C' denotes the weight compression rate.
    }
    \label{tab:resnet_50_imagenet}
\end{table*}
\begin{table*}[t!]
       \small
    \begin{tabular}{cccccccccc} 
        \toprule
        Method & W-bit & A-bit & Top-1/Full & Top-1/Quant & Top-1/Drop &  W-C & Searching Data \\
        \midrule
         DC \cite{Han2015DeepCC} & MP & 32 & 71.87 & 58.07 & -13.80 & 13.93$\times$ &  - \\
         HAQ \cite{wang2019haqhardwareawareautomatedquantization} & MP & 32 & 71.87 & 66.75 & -5.12 & \textbf{14.07$\times$} & -  \\
         MPQCO \cite{chen2021mixedprecisionquantizationneuralnetworks} & MP & 8 & 71.88 & 68.52 & -3.36 & 13.99$\times$ & \textbf{1024}\\
        \cdashline{1-8}    
        \textbf{InfoQ (Ours)} & \textbf{MP}& \textbf{8}& \textbf{72.02} & \textbf{69.83} & \textbf{-2.19} & 14.00$\times$ & \textbf{1024} \\
        \bottomrule
    \end{tabular}
     \centering
    \caption{Results for MobileNetv2 on ImageNet under weight compression constraints, denoted by 'W-C'.
    }
    \label{tab:mobilenetv2_imagenet}
\end{table*}


\section{Experiments}
We conduct a comprehensive evaluation of the \textit{InfoQ} framework across three dimensions. First, we present a state-of-the-art (SOTA) comparison on standard benchmarks after quantization-aware training (QAT). Second, we analyze the post-quantization accuracy \textit{without} QAT to directly assess the quality of our sensitivity metric \cite{Deng2023MixedPrecisionQF}. Finally, we evaluate the search efficiency of our method.

\subsection{State-of-the-Art Comparison with QAT}

To evaluate the effectiveness of the bit configurations found by \textit{InfoQ}, we follow standard community practices. We perform experiments on the ImageNet (ILSVRC12) dataset \cite{5206848} using three widely-used architectures: ResNet18/50 \cite{he2015deepresiduallearningimage} and MobileNetV2 \cite{sandler2019mobilenetv2invertedresidualslinear}. For a given model and hardware constraint (e.g., model size or BitOps), we first use \textit{InfoQ} to determine the optimal bit-width configuration. We then apply QAT using the Learned Step-Size (LSQ) method \cite{Esser2019LearnedSS} to recover accuracy. The candidate bit-width set for both weights and activations is $\boldsymbol{B} = \{2, 3, 4, 5, 6, 7, 8\}$. We compare \textit{InfoQ} against a wide range of leading uniform-precision and mixed-precision methods.

As shown in Table~\ref{tab:resnet_18_imagenet}, \textit{InfoQ} demonstrates superior performance on ResNet18. In a weight-only quantization setting (avg. 3-bit weights, 8-bit activations), our method achieves an accuracy of 70.94\%, surpassing the full-precision baseline and outperforming other methods like OMPQ and MPQCO. In the more challenging setting of joint weight and activation quantization under a 23.04 G-BitOps constraint, \textit{InfoQ} achieves 69.99\% top-1 accuracy, marking the smallest accuracy degradation (-0.61\%) among all listed methods. This result is particularly strong when compared to LIMPQ, which reports a similar BitOps budget but achieves it by leaving the first layer activations at 32-bit, a significant deviation from a fully quantized model.

On the larger ResNet50 model, under a strict 12.2$\times$ weight compression constraint, \textit{InfoQ} achieves a state-of-the-art top-1 accuracy of 77.03\% (Table~\ref{tab:resnet_50_imagenet}). This result represents an accuracy drop of only -0.34\% from its full-precision baseline, outperforming prominent methods like HAWQv2 (-1.50\%) and LIMPQ (-0.60\%). 
Furthermore, this result is achieved with a training-free search phase that uses only $6000$ calibration samples, in contrast to retraining-based proxies like LIMPQ which require a substantial fraction of the full training set.

To evaluate performance on a non-residual, depthwise-separable architecture, we test on MobileNetV2. We set an aggressive 14.00$\times$ weight compression target to ensure quantization is applied across the entire network, not just the final fully-connected layer. As shown in Table~\ref{tab:mobilenetv2_imagenet}, \textit{InfoQ} excels in this setting, achieving 69.83\% accuracy. This significantly outperforms prior methods such as HAQ (-5.12\% drop) and MPQCO (-3.36\% drop) at a comparable compression rate.

\begin{algorithm}[t!]
\caption{InfoQ Sensitivity Score Computation}
    \label{alg:smi_computation}
    \textbf{Input}: Pre-trained model $f$, calibration data $(\boldsymbol{X_c}, \boldsymbol{Y_c})$, candidate bit-widths $\boldsymbol{B}$, observer layers sets $\mathcal{O}_{XL}, \mathcal{O}_{LY}$.
    \\
    \textbf{Output}: Sensitivity scores $S(\ell, b)$ for each layer $\ell$ and bit-width $b$.
    \begin{algorithmic}[1] 
    \STATE {\# \textit{1. Compute Baseline Information on 8-bit model}}
    \STATE Let $\boldsymbol{s}_{\text{8bit}}$ be the configuration with all layers at 8-bit.
    \STATE Get baseline activations $\{\boldsymbol{L}_{j, \text{8bit}}\}_{j \in \mathcal{O}_{XL} \cup \mathcal{O}_{LY}}$ from $f(\boldsymbol{X_c}; \boldsymbol{s}_{\text{8bit}})$.
        \STATE $I_{X,L}^{(j, \text{8bit})} \leftarrow \text{SMI}(\boldsymbol{X_c}, \boldsymbol{L}_{j, \text{8bit}})$ for observers $j \in \mathcal{O}_{XL}$
        \STATE $I_{L,Y}^{(j, \text{8bit})} \leftarrow \text{SMI}(\boldsymbol{L}_{j, \text{8bit}}, \boldsymbol{Y_c})$ for observers $j \in \mathcal{O}_{LY}$
    \STATE {\# \textit{2. Compute Sensitivity for each (layer, bit-width) pair}}
    \FOR{each quantizable layer $\ell = 1, \dots, L$}
        \FOR{each bit-width $b \in \boldsymbol{B}$}
            \STATE {\# \textit{Perturb model by quantizing only layer $\ell$ to bit $b$}}
            \STATE $\boldsymbol{s}_{\text{pert}} \leftarrow \boldsymbol{s}_{\text{8bit}}$
            \STATE Set layer $\ell$ to bit-width $b$ in $\boldsymbol{s}_{\text{pert}}$.
            \STATE Get perturbed activations $\{\boldsymbol{L}_{j, b}\}_{j > \ell}$ from $f(\boldsymbol{X_c}; \boldsymbol{s}_{\text{pert}})$.
            \STATE {\# \textit{Calculate normalized score as per Eq.~\ref{eq:score}}}
            \STATE $S(\ell, b, \mathcal{O}_{XL}^{j>i}, \mathcal{O}_{LY}^{j>i}) \leftarrow$ Eq.~\eqref{eq:score}
%
        \ENDFOR
    \ENDFOR
    \STATE \textbf{return} $S$
\end{algorithmic}
\end{algorithm}

\subsection{Direct Evaluation of the Sensitivity Metric}
The performance of a model after QAT depends on both the quality of the bit-width configuration and the effectiveness of the fine-tuning process. To isolate and directly evaluate the quality of the bit allocation produced by our sensitivity metric, we compare its performance \textit{before} any QAT, a setup often referred to as post-training quantization (PTQ) \cite{Deng2023MixedPrecisionQF}. A superior PTQ accuracy indicates that the underlying sensitivity metric is more effective at identifying a robust quantization configuration.

\begin{figure}[t!]
    \centering
    \begin{subfigure}{0.48\columnwidth}
        \includegraphics[width=\columnwidth]{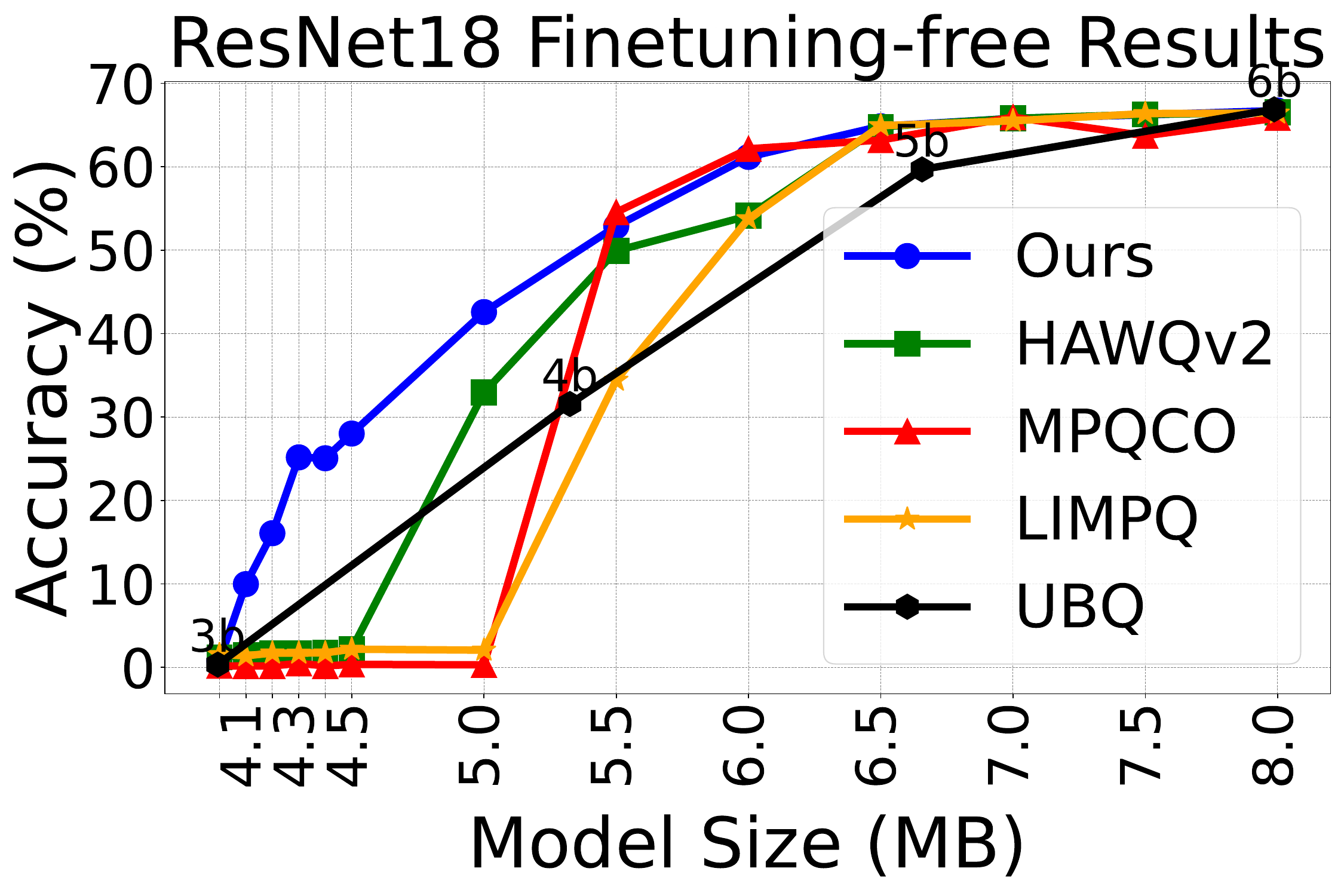}
        \label{}
    \end{subfigure}
    \begin{subfigure}{0.48\columnwidth}
        \includegraphics[width=\columnwidth]{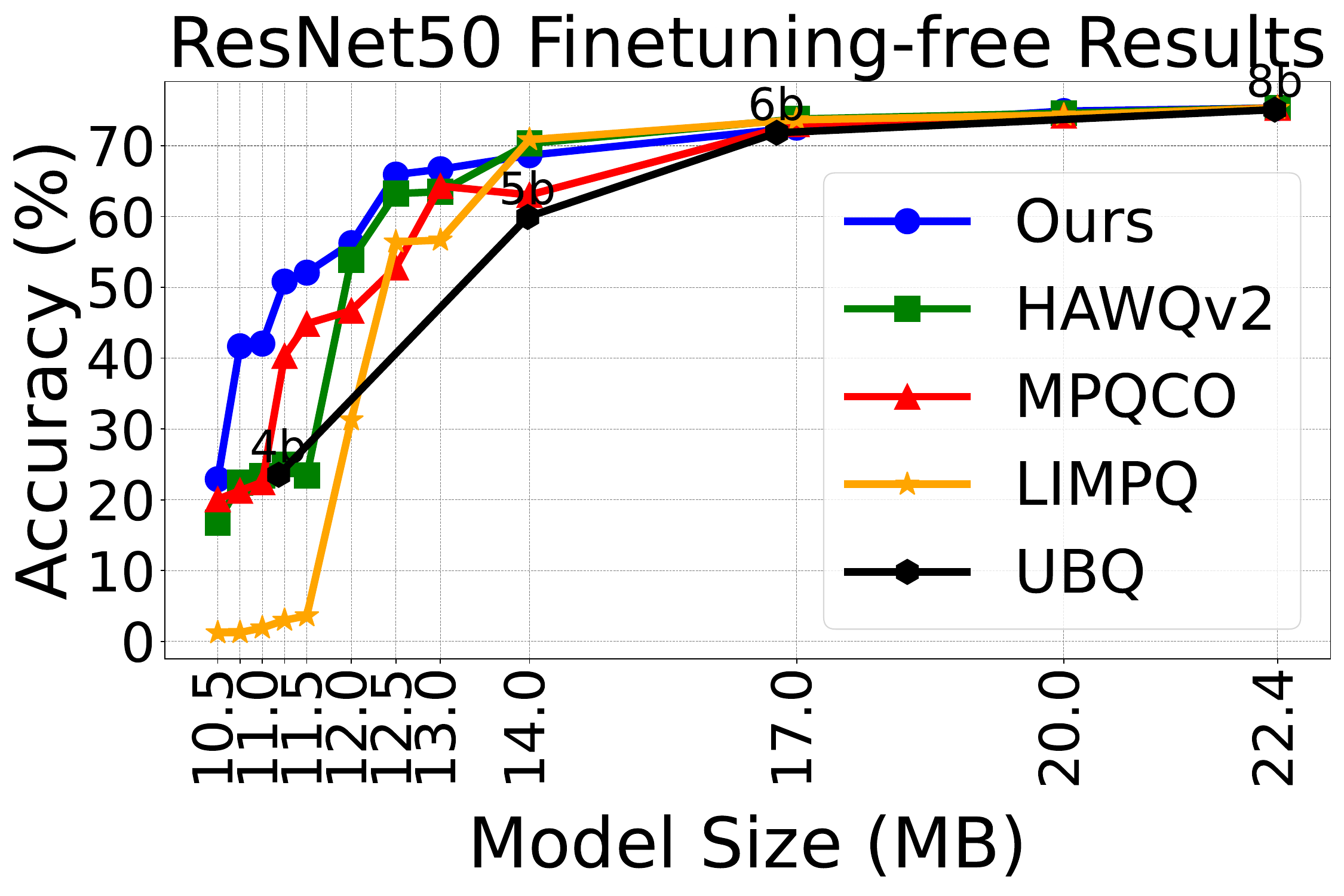}
        \label{}
    \end{subfigure}

    \label{fig:results_without_finetuning}
        \begin{subfigure}{0.48\columnwidth}
        \includegraphics[width=\columnwidth]{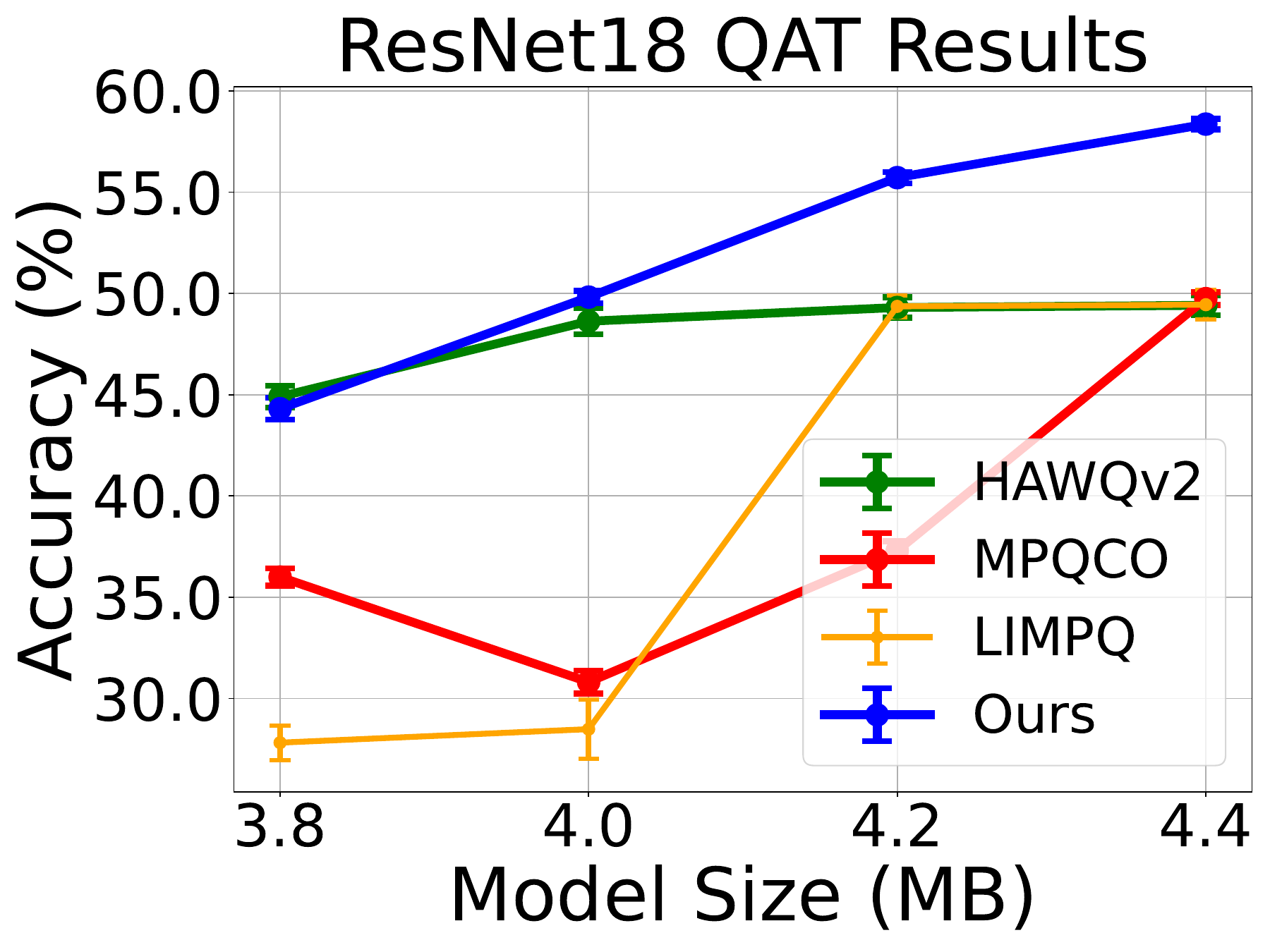}
        \label{}
    \end{subfigure}
    \begin{subfigure}{0.48\columnwidth}
        \includegraphics[width=\columnwidth]{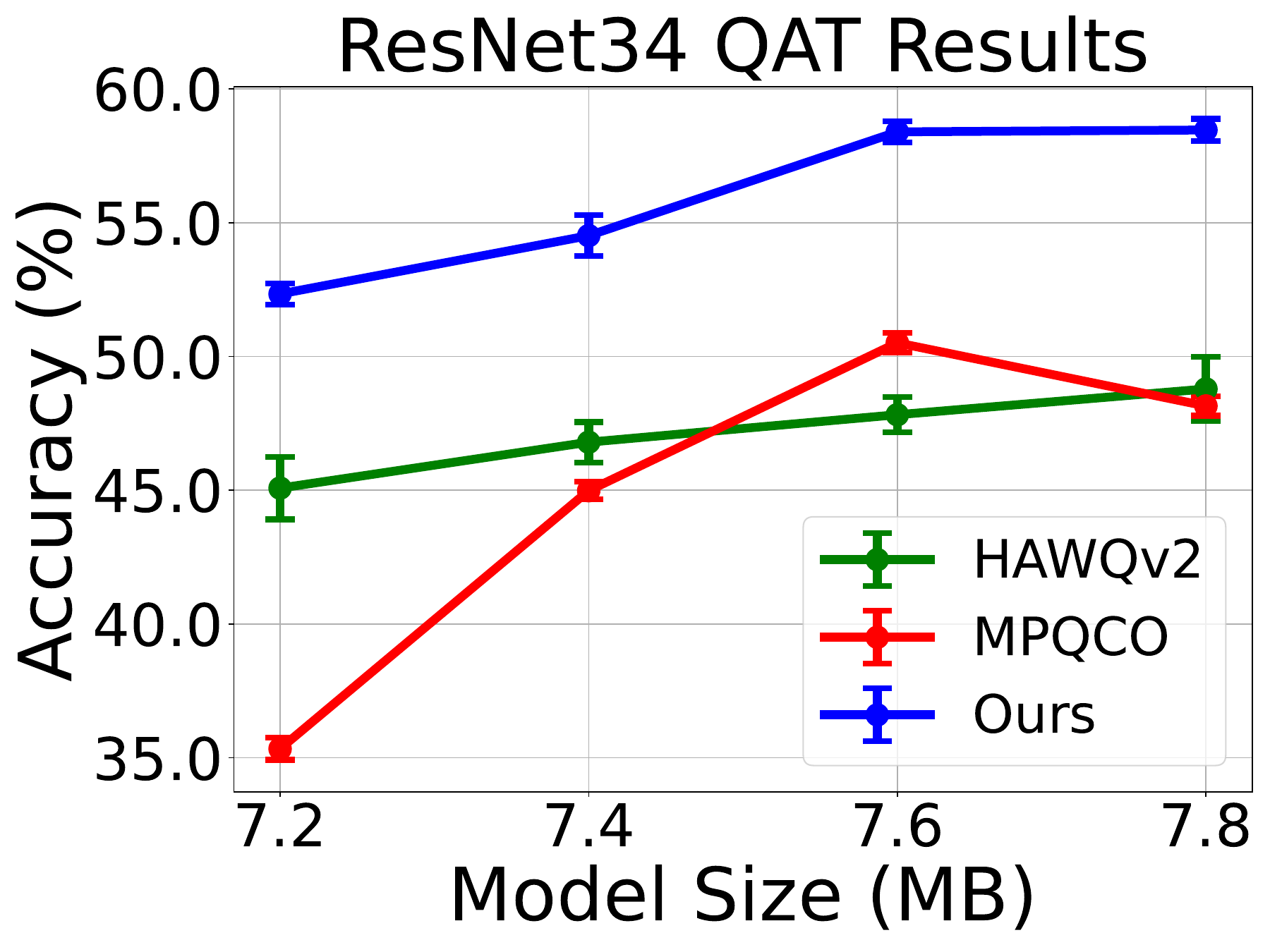}
        \label{}
    \end{subfigure}
    \caption{Direct evaluation of our method against state-of-the-art \textit{criterion-based} methods on ResNet18 and ResNet50 (top). The results are derived from best-performing bit-width configurations according to sensitivity scores for each parameter size constraint. QAT results (bottom) are averaged over 25 runs, each conducted for 5 epochs with 25,000 samples. LIMPQ source code does not provide search configurations for ResNet34.
    }
    \label{fig:results_with_finetuning}
\end{figure}

We compare \textit{InfoQ} against leading criterion-based methods (HAWQv2, MPQCO, LIMPQ) across a range of model size constraints for ResNet18/34/50 on ImageNet. For each method and each constraint, we generate the optimal bit-width configuration and report the corresponding PTQ accuracy.
The results are presented in Figure~\ref{fig:results_with_finetuning}. In high-compression regimes (e.g., model sizes approaching that of uniform 3-bit quantization), the performance gap between methods becomes most apparent. In these scenarios, \textit{InfoQ} consistently outperforms all other methods, often by a significant margin. For instance, on ResNet18 at a 4.5MB constraint, \textit{InfoQ} yields a configuration that is 25\% more accurate than its closest competitor. 
This demonstrates that our global, information-based sensitivity metric is fundamentally more effective at preserving model accuracy than local, Hessian-based or retraining-based proxies.
Furthermore, a superior initial configuration provides a better starting point for fine-tuning. As shown in Figure~\ref{fig:results_with_finetuning}, we observe that bit-width assignments from \textit{InfoQ} also lead to higher accuracy after a short QAT schedule. This confirms that the benefits of our more accurate sensitivity metric compound during the fine-tuning process.

\begin{figure}[t!]
    \centering
    \begin{subfigure}{0.95\columnwidth}
        \includegraphics[width=\columnwidth]{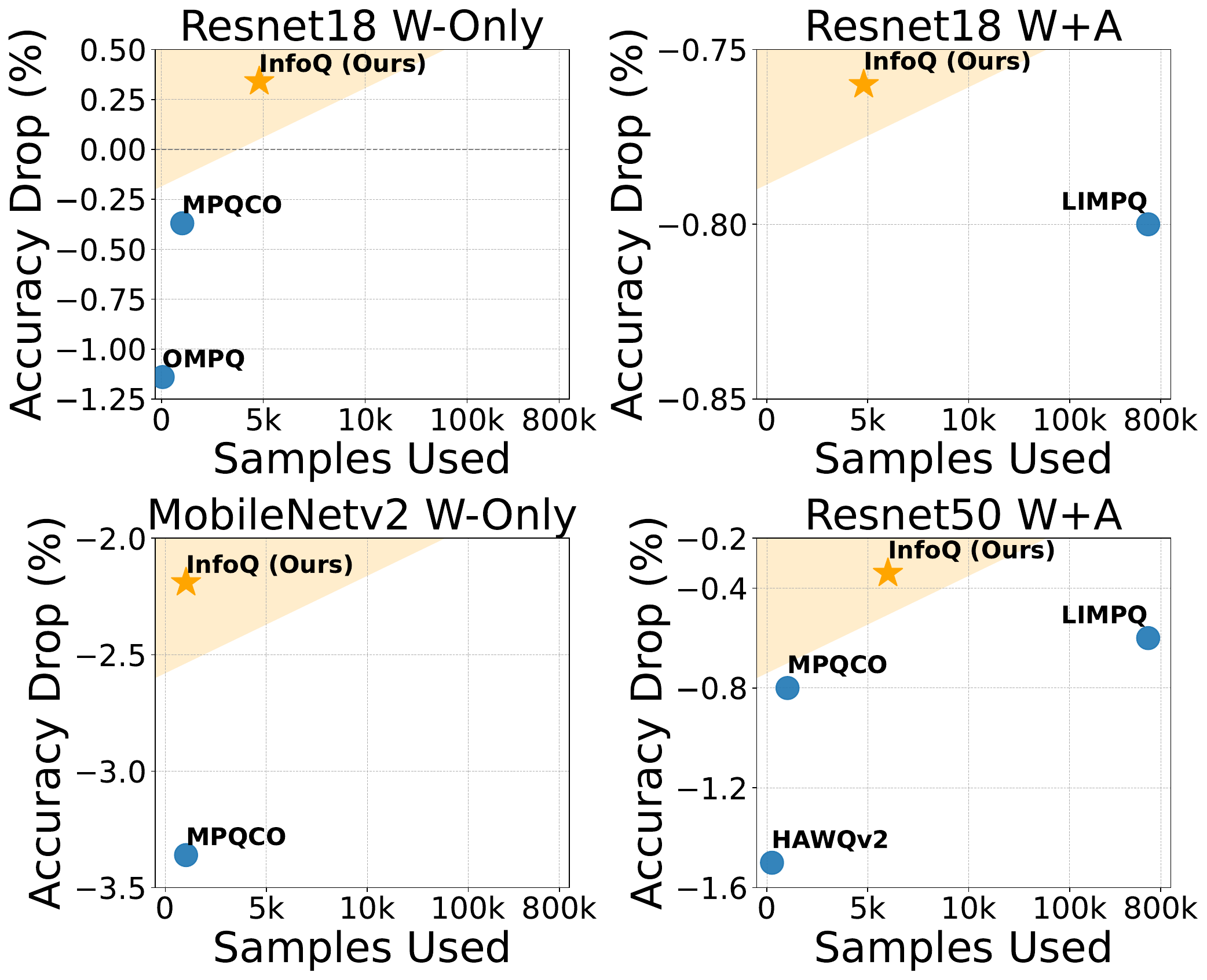}
        \label{}
    \end{subfigure}
    \caption{Performance of our method against state-of-the-art \textit{criterion-based} methods. 'W-Only' and 'W+A' mean 'weight-only' and 'weight+activation' MPQ.
    }
    \label{fig:sota_comp}
\end{figure}
\subsection{Search Efficiency Analysis}
A key advantage of \textit{InfoQ} is its high search efficiency. The cost is composed of two phases: a one-time sensitivity analysis and the near-instantaneous bit allocation.

\paragraph{Sensitivity Analysis Cost.} 
The primary computational cost of our method is a one-time, training-free analysis. For a network with $L$ quantizable layers and a set of candidate bit-widths $\boldsymbol{B}$, the process requires $L \times |\boldsymbol{B}|$ forward passes over a small, labeled calibration dataset to collect the necessary activations. 
From these activations, we compute SMI estimates at the pre-selected observer layers. This entire analysis is backpropagation-free and highly parallelizable.

The selection of observer layers is performed via a correlation analysis on a small subset of the calibration data. Importantly, the values from this step can be reused in the subsequent sensitivity analysis (Algorithm~\ref{alg:smi_computation}, Eq.~\ref{eq:score}), making its computational cost negligible.

Retraining-based proxy methods like LIMPQ \cite{tang2023mixedprecisionneuralnetworkquantization} require extensive training on a large fraction of the dataset (e.g., ~600,000 samples for ImageNet) to learn their sensitivity indicators. 
In contrast, Hessian-based methods like HAWQv2 are faster but, as we have shown, yield suboptimal bit assignments due to their local scope. \textit{InfoQ} thus occupies a unique and highly favorable position in the accuracy-efficiency trade-off, achieving SOTA accuracy with a practical, one-shot analysis, as shown in the Figure \ref{fig:sota_comp}.

\paragraph{Bit Allocation Cost.} 
Once the sensitivity scores $S(\ell,b)$ are pre-computed, the bit allocation phase is extremely fast. For any given hardware constraint (e.g., a target model size or BitOps), the optimal bit-width configuration is found by solving the ILP problem (Eq.~\ref{eq:ilp}). This is solved by an off-the-shelf solver in milliseconds. A significant practical benefit of this decoupling is that the same set of pre-computed sensitivity scores can be reused instantly to generate optimal configurations for multiple different hardware targets or budget constraints, without re-running the more expensive sensitivity analysis.

\section{Conclusion}
In this work, we addressed a fundamental limitation of existing criterion-based MPQ methods: their reliance on local sensitivity metrics. We argued that the true impact of quantizing a layer is a global phenomenon, best measured by the resulting disruption to the network information flow.
To this end, we introduced \textit{InfoQ}, a novel framework that is based on this principle. By using Sliced Mutual Information to directly quantify the cascading effects of quantization on downstream layers, \textit{InfoQ} constructs an accurate, global sensitivity metric. This metric enables the determination of optimal bit-width configurations through a fast, training-free analysis, followed by an instantaneous ILP-based allocation.

Our extensive experiments on ImageNet demonstrate that this information-theoretic approach yields state-of-the-art results, outperforming both Hessian-based and retraining-based methods on ResNet and MobileNetV2 architectures, particularly in high-compression regimes. Our work establishes a practical methodology for applying information-theoretic diagnostics to model compression, providing a principled foundation for bit-width (and more in general, perturbation robustness) allocation in deep neural network layers.

\bibliography{reference}

\clearpage
\appendix

\section{Observer Layer Selection Methodology}
\label{app:observer_selection}

This section details the empirical procedure used to identify the optimal set of \textit{observer layers} for a given network architecture. The goal is to select layers where the measured change in Sliced Mutual Information (SMI) is most predictive of the final degradation in task performance.

\subsection{Correlation Analysis Procedure}
For a given network architecture, we generate a dataset to correlate informational changes $\Delta{\text{SMI}}$ with accuracy degradation. The procedure is reported in Algorithm~\ref{alg:obs_select}.
For each quantizable layer $i$ in the network, we first create a perturbed model by quantizing only layer $i$ to a low bit-width (e.g., 2-bit), while all other layers remain at a baseline 8-bit precision. Then, we perform a forward pass over a validation set and record the final top-1 accuracy degradation,~$\Delta\text{Acc}$. For every subsequent layer $j > i$, we compute the informational changes $\Delta\text{SMI}_{X,L}^{(i,j)}$ and $\Delta\text{SMI}_{L,Y}^{(i,j)}$ as per Eqs.~\ref{eq:deltaSMI} in the main text. This process yields, for each layer $j$, a set of paired observations $(\Delta\text{SMI}^{(j)}, \Delta\text{Acc})$. We then compute the Pearson correlation coefficient, $\rho$, between these two variables for each layer and for each type of SMI. A strong negative correlation (e.g., $|\rho| > -0.70$) signifies that the SMI change at that layer is a highly predictive proxy for performance loss. Moreover, we focus on the layers at the end of each block for simplicity, which enables us to analyze observers block by block.

\begin{algorithm}[t!]
\caption{Data-Driven Observer layer Selection}
\label{alg:observer_selection}
\textbf{Input}: Pre-trained model $f$, validation dataset $(\boldsymbol{X_v}, \boldsymbol{Y_v})$, low bit-width $b_{\text{low}}$ (e.g., 2), correlation threshold $\tau$ (e.g., 0.7).
\\
\textbf{Output}: Observer layer sets $\mathcal{O}_{XL}$, $\mathcal{O}_{LY}$.

\begin{algorithmic}[1] 
    \STATE \textcolor{black}{\# \textit{1. Generate a dataset of (Accuracy Drop, $\Delta$SMI Change) pairs}}
    \STATE Initialize data lists: $D_{XL}[j] \leftarrow []$, $D_{LY}[j] \leftarrow []$ for each layer $j$.
    \STATE Acc$_{\text{8bit}} \leftarrow \text{Evaluate}(f, \boldsymbol{X_v}, \boldsymbol{Y_v})$ with all layers at 8-bit.
    \STATE Compute baseline SMI values $I_{X,L}^{(j, \text{8bit})}$, $I_{L,Y}^{(j, \text{8bit})}$ for all layers $j$.
    
    \FOR{each quantizable layer $i = 1, \dots, L$}
        \STATE \textcolor{black}{\# \textit{Perturb layer i and measure the global impact}}
        \STATE Let $\boldsymbol{s}_{\text{pert}}$ be a config with layer $i$ at $b_{\text{low}}$ and others at~8-bit.
        \STATE Acc$_{\text{pert}} \leftarrow \text{Evaluate}(f(\cdot ; \boldsymbol{s}_{\text{pert}}), \boldsymbol{X_v}, \boldsymbol{Y_v})$.
        \STATE $\Delta\text{Acc} \leftarrow \text{Acc}_{\text{8bit}} - \text{Acc}_{\text{pert}}$.

        \FOR{each layer $j$ where $j > i$}
            \STATE Compute perturbed SMI values $I_{X,L}^{(j, b_{\text{low}})}$, $I_{L,Y}^{(j, b_{\text{low}})}$.
            \STATE $\Delta I_{X,L} \leftarrow |I_{X,L}^{(j, \text{8bit})} - I_{X,L}^{(j, b_{\text{low}})}|$.
            \STATE $\Delta I_{L,Y} \leftarrow |I_{L,Y}^{(j, \text{8bit})} - I_{L,Y}^{(j, b_{\text{low}})}|$.
            \STATE Append $(\Delta I_{X,L}, \Delta\text{Acc})$ to $D_{XL}[j]$.
            \STATE Append $(\Delta I_{L,Y}, \Delta\text{Acc})$ to $D_{LY}[j]$.
        \ENDFOR
    \ENDFOR

    \STATE
    \STATE \textcolor{black}{\# \textit{2. Select observers based on correlation}}
    \STATE $\mathcal{O}_{XL} \leftarrow \emptyset$, $\mathcal{O}_{LY} \leftarrow \emptyset$.
    \FOR{each layer $j = 1, \dots, L$}
            \STATE Let $(\boldsymbol{\Delta I_{XL}}, \boldsymbol{\Delta\text{Acc}}) = D_{XL}[j]$.
            \STATE $\rho_{XL} \leftarrow \text{PearsonCorr}(\boldsymbol{\Delta I_{XL}}, \boldsymbol{\Delta\text{Acc}})$.
            \IF{$|\rho_{XL}| > \tau$}
                \STATE $\mathcal{O}_{XL} \leftarrow \mathcal{O}_{XL} \cup \{j\}$.
            \ENDIF
    \ENDFOR
    \STATE \textcolor{black}{\# \textit{3. Select observers starting from the last block, stop when $\rho$ is below threshold. }}
    \FOR{each layer $j = L,L-1 \dots, 1$}
            \STATE Let $(\boldsymbol{\Delta I_{LY}}, \boldsymbol{\Delta\text{Acc}}) = D_{LY}[j]$.
            \STATE $\rho_{LY} \leftarrow \text{PearsonCorr}(\boldsymbol{\Delta I_{LY}}, \boldsymbol{\Delta\text{Acc}})$.
            \IF{$|\rho_{LY}| > \tau$}
                \STATE $\mathcal{O}_{LY} \leftarrow \mathcal{O}_{LY} \cup \{j\}$.
            \ELSE
                \STATE break
            \ENDIF        
    \ENDFOR
    \STATE \textbf{return} $\mathcal{O}_{XL}, \mathcal{O}_{LY}$
\end{algorithmic}
\label{alg:obs_select}
\end{algorithm}

\begin{figure*}[htbp]
    \centering
    \includegraphics[width=1\linewidth]{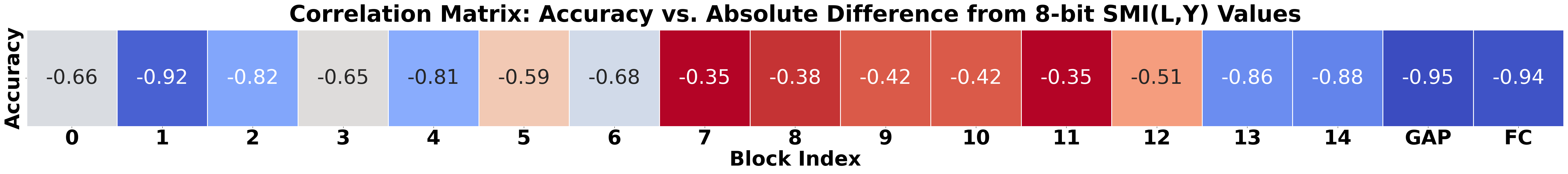}
        \caption{Correlation matrix for ResNet50, showing the relationship between accuracy degradation and the change in target-relevant information, $\Delta SMI(L,Y)$, measured at the output of each layer. Note the strong negative correlation in the final layers of the network.}
    \label{fig:corr_ly}
\end{figure*}
\begin{figure*}[htbp]
    \centering
    \includegraphics[width=0.7\linewidth]{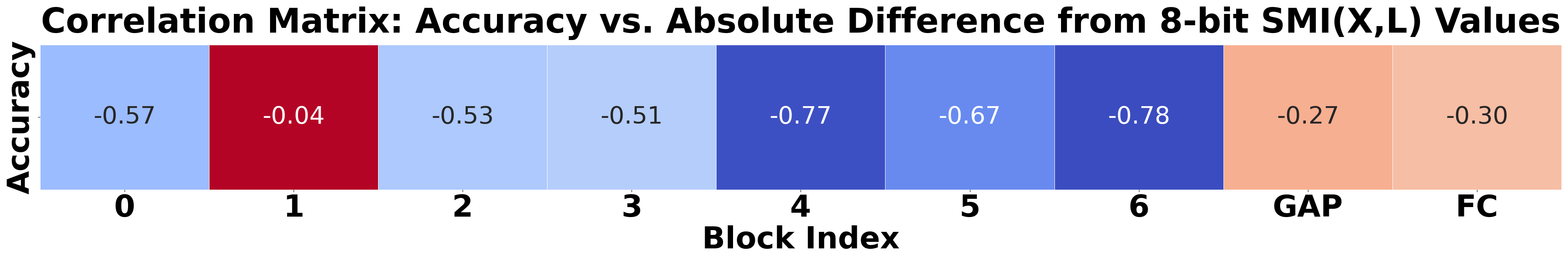}
        \caption{Correlation matrix for ResNet18, showing the relationship between accuracy degradation and the change in input-relevant information, $\Delta SMI(X,L)$, measured at the output of each layer. The strongest correlations are found in the intermediate layers.}
    \label{fig:corr_xl}
\end{figure*}

\subsection{Empirical Results and Final Selection}
We present representative correlation matrices for ResNet architectures in Figure~\ref{fig:corr_ly} and Figure~\ref{fig:corr_xl}. 
Concerning observers for $\Delta\text{SMI}(L,Y)$,  the correlation is consistently strongest in the final classification layers. For ResNet50 (Figure~\ref{fig:corr_ly}), the correlation at the \texttt{GlobalAvgPool} and \texttt{FC} layers approaches $\rho = -0.95$. Any loss of task-relevant information in these final stages has a direct and predictable negative impact on accuracy.
Concerning observers for $\Delta\text{SMI}(X,L)$, the strongest correlation is found in the intermediate feature-extraction layers. For ResNet18 (Figure~\ref{fig:corr_xl}), this occurs in layers 4-6. The final layers, designed to discard input-irrelevant information, show a much weaker correlation, as expected.

It is important to note a limitation of this analysis for the earliest layers of the network. The number of data points available to compute the correlation for a layer~$j$ is limited by the number of preceding layers~$i < j$. For the very first layers, this sample size is too small to yield a statistically reliable correlation coefficient. 
Therefore, the earliest layers are not selected as observers due to an insufficient number of preceding layers to generate a statistically reliable correlation.  We show that the correlation between $\Delta\text{SMI}(L,Y)$ and $\Delta\text{Acc}$ is higher for deeper layers in a neural network.  Consequently, our method selects observers starting from the final layer and working backward, stopping when the correlation drops below a specified threshold. This approach avoids the aforementioned limitation of earliest layers.
Our objective is to select a set of highly correlated observer layers to ensure the sensitivity analysis remains efficient. 

Based on the correlation analysis, we select the final sets of observer layers $\mathcal{O}_{XL}$ and $\mathcal{O}_{LY}$ for our experiments.
The set of observers $\mathcal{O}_{LY}$ are mostly the final pooling, fully-connected or last few convolutional layers that exhibit a correlation $|\rho| > 0.70$ for $\Delta\text{SMI}(L,Y)$.
Similarly, the set of observers $\mathcal{O}_{XL}$ generally includes the layers in the intermediate stages (blocks until the final pooling layer) that exhibit a correlation $|\rho| > 0.70$ for $\Delta\text{SMI}(X,L)$.
This data-driven procedure grounds our sensitivity metric in the specific information flow of each architecture. 

\begin{figure}
    \centering
    \includegraphics[width=\linewidth]{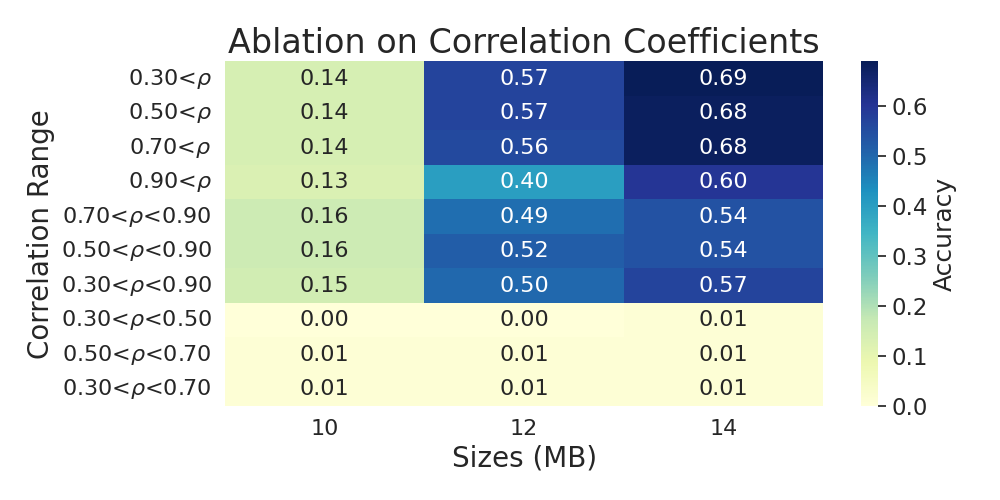}
    \caption{PTQ accuracies of different correlation coefficient range selections under various model size constraints.}
    \label{fig:corr_matrix}
\end{figure}

\begin{figure}
    \centering
    \includegraphics[width=\linewidth]{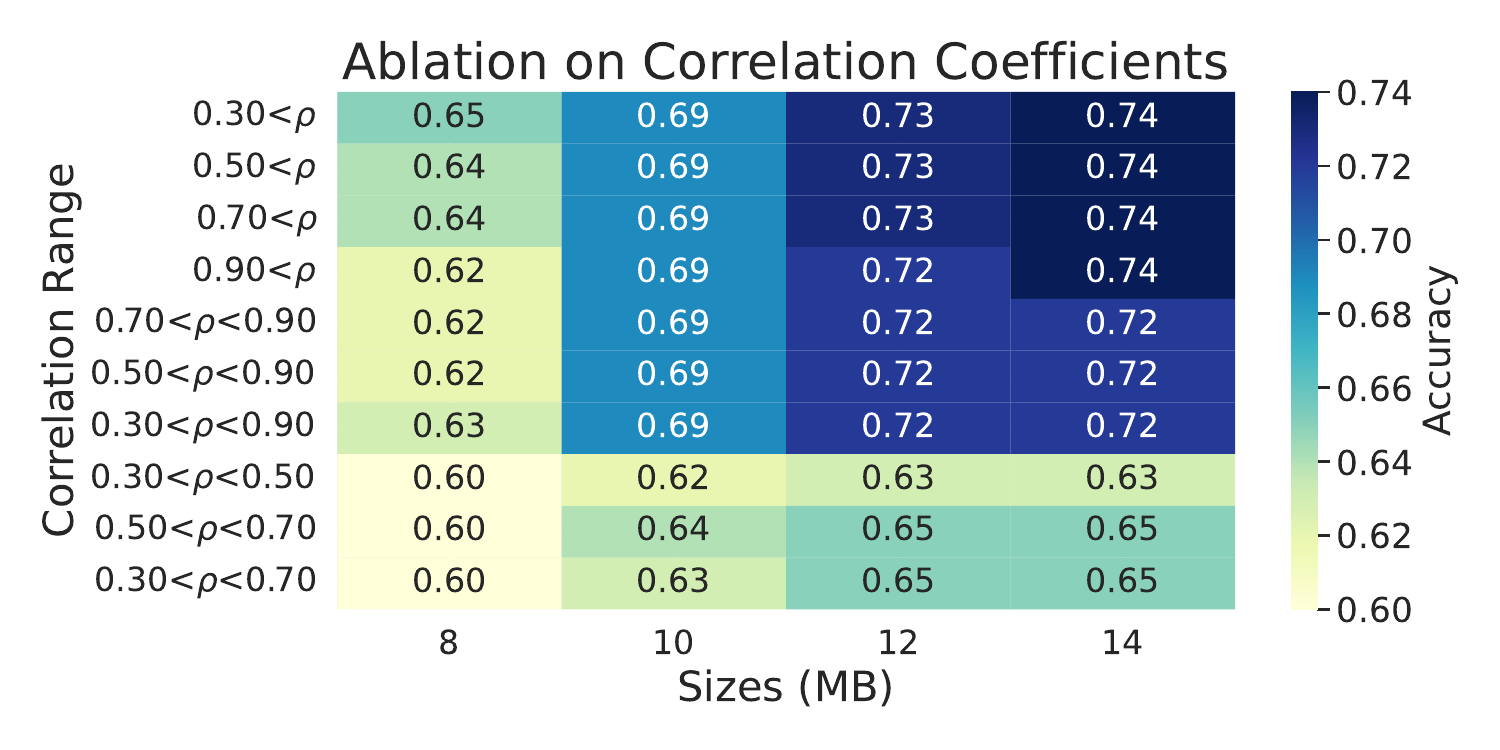}
    \caption{QAT accuracies of different correlation coefficient range selections under various model size constraints. Results are averaged over training runs of 1 epoch training with 5k samples. Standard deviations are not annotated as they are $<0.001$ for all cases.}
    \label{fig:corr_matrix_qat}
\end{figure}

\begin{figure}[t]
    \centering
    \includegraphics[width=0.95\linewidth]{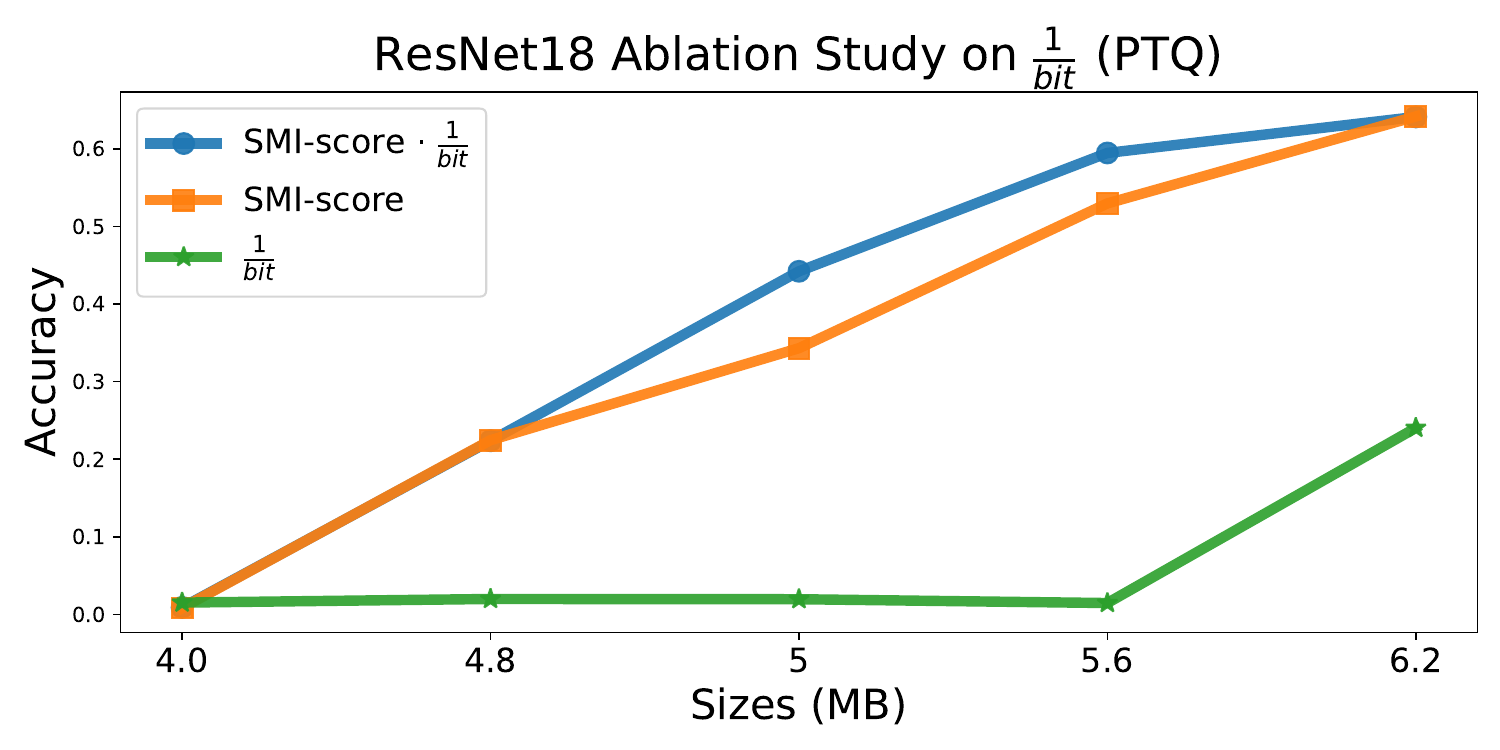}
\caption{ResNet18 Post-Training Quantization (PTQ) accuracy for different model sizes. The plot compares our full sensitivity score (SMI-score $\cdot \frac{1}{b}$), the score without the penalty term (SMI-score), and a baseline using only the penalty term ($\frac{1}{b}$).}

    \label{fig:ptq_bit}
\end{figure}

\begin{figure}[t]
    \centering
    \includegraphics[width=0.95\linewidth]{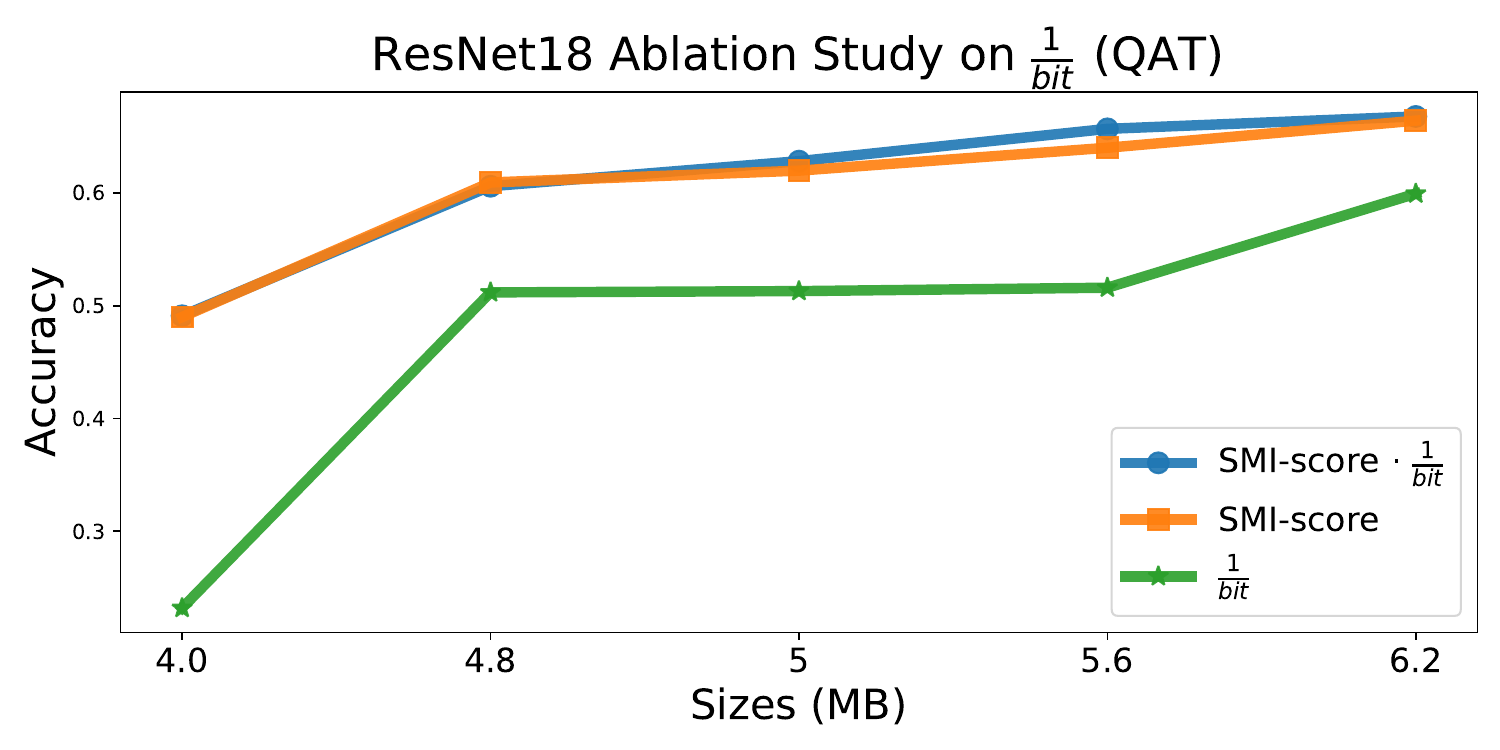}
\caption{ResNet18 Quantization-Aware Training (QAT) accuracy for different model sizes. The comparison confirms the trends observed in the PTQ setting, where the full sensitivity score performs best and the information-theoretic component is dominant.}

    \label{fig:qat_bit}
\end{figure}

\subsection{Ablation Study on Observer Selection Threshold}
\label{app:ablation_threshold}

To empirically validate our observer selection strategy, we conduct an ablation study on the correlation threshold, $\tau$. Our methodology selects observer layers where the Pearson correlation coefficient, $\rho$, between the measured $\Delta$SMI and the final accuracy drop exceeds this threshold ($|\rho| > \tau$). This study investigates the impact of varying $\tau$ on the final model accuracy, thereby justifying our choice of $\tau=0.7$.

We evaluate this on ResNet50 by generating bit-width allocations using different sets of observer layers, selected based on varying correlation ranges. The final PTQ and QAT accuracies for these configurations are presented as heatmaps in Figure~\ref{fig:corr_matrix} and Figure~\ref{fig:corr_matrix_qat}, respectively, across several model size constraints.


Relying solely on layers with a weak correlation to accuracy degradation is catastrophic. As shown in the bottom rows of the heatmaps (e.g., $0.3 < |\rho| < 0.5$), the resulting bit-allocations lead to near-zero PTQ accuracy and severely suboptimal QAT results. This confirms our central premise: if the informational change at an observer is not a strong predictor of performance loss, it is an ineffective proxy for quantization sensitivity.

The results validate $|\rho| > 0.7$ as a robust and effective threshold. The row corresponding to $|\rho| > 0.7$ achieves high accuracy across all budgets (e.g., 0.68 PTQ accuracy at 14MB). Crucially, expanding the observer set to include less-correlated layers (e.g., by using a threshold of $|\rho| > 0.5$) yields nearly identical performance. This indicates that the sensitivity calculation is dominated by the strong signal from the highly-correlated observers ($|\rho| > 0.7$), making the final bit-allocation robust to the inclusion of weaker, less-informative signals. Therefore, selecting all layers where $|\rho| > 0.7$ is sufficient.

While a high correlation is good, an excessively stringent threshold like $|\rho| > 0.9$ can be suboptimal. The heatmaps show that this selection, while still effective, yields consistently lower accuracy than our chosen threshold of $|\rho| > 0.7$ (e.g., 0.60 vs. 0.68 PTQ accuracy at 14MB). A likely explanation is that such a high threshold drastically reduces the number of observers (to just two final layers for ResNet50). While highly predictive, this small set may provide a less comprehensive view of the global information degradation compared to the slightly larger set captured by $|\rho| > 0.7$.

This study provides strong empirical justification for using a correlation threshold of $\tau=0.7$. This value is stringent enough to filter out uninformative layers that would introduce noise into the sensitivity metric, yet not so restrictive that it risks an incomplete assessment of the network global information flow. This choice ensures that the InfoQ sensitivity score is both robust and maximally effective.

\section{Penalty Term $\frac{1}{b}$}
Our sensitivity scoring function, as defined in Eq. 4, includes a $\frac{1}{b}$ penalty term, where $b$ is the bit-width. This term is designed to penalize lower bit-widths, thereby promoting more stable quantization configurations by favoring higher bit-widths when two options yield a comparable change in Sliced Mutual Information (SMI). To isolate and verify the contribution of this component, we conduct a comprehensive ablation study on the ResNet18 architecture evaluated on ImageNet.

We evaluate three distinct methods for generating bit-width assignments across various model size constraints:
(1) The complete InfoQ sensitivity metric (SMI-score $\cdot \frac{1}{b}$), which combines the information-theoretic score with the bit-width penalty; (2) The sensitivity metric based solely on the normalized change in SMI (SMI-score), with the penalty term~$1/b$ removed; (3) A simple heuristic that assigns bit-widths based only on the penalty term ($\frac{1}{b}$), completely discarding the information-theoretic metric.

The performance of these three methods under Post-Training Quantization (PTQ) is presented in Figure~\ref{fig:ptq_bit}. The results show that the full scoring function (SMI-score $\cdot \frac{1}{b}$) achieves slightly higher accuracy than using the SMI-score alone, validating the penalty term role in refining the bit allocation. Relying solely on the $\frac{1}{b}$ heuristic leads to a catastrophic drop in performance, yielding near-zero accuracy for most model sizes. This demonstrates that while the penalty term is a useful component, it is insufficient on its own.

We extend this analysis to a Quantization-Aware Training (QAT) setting, with the results shown in Figure~\ref{fig:qat_bit}. While QAT significantly improves the performance of all methods, the relative hierarchy remains consistent. The full InfoQ method continues to hold a slight advantage over the SMI-score without the penalty. Although the performance of the $\frac{1}{b}$ heuristic is substantially recovered through training, it remains markedly inferior to the information-theoretic approaches, especially in more aggressive compression regimes (i.e., smaller model sizes).

In conclusion, this ablation study confirms two key points. First, the $\frac{1}{b}$ penalty term provides a small but consistent improvement to our method performance. Second, and more importantly, the dramatic failure of the penalty-only heuristic underscores the fundamental superiority and critical importance of our global, information-theoretic sensitivity metric for effective mixed-precision quantization.

\section{Bit Assignments}
\begin{figure*}[htbp]
    \centering
    \includegraphics[width=0.5\linewidth]{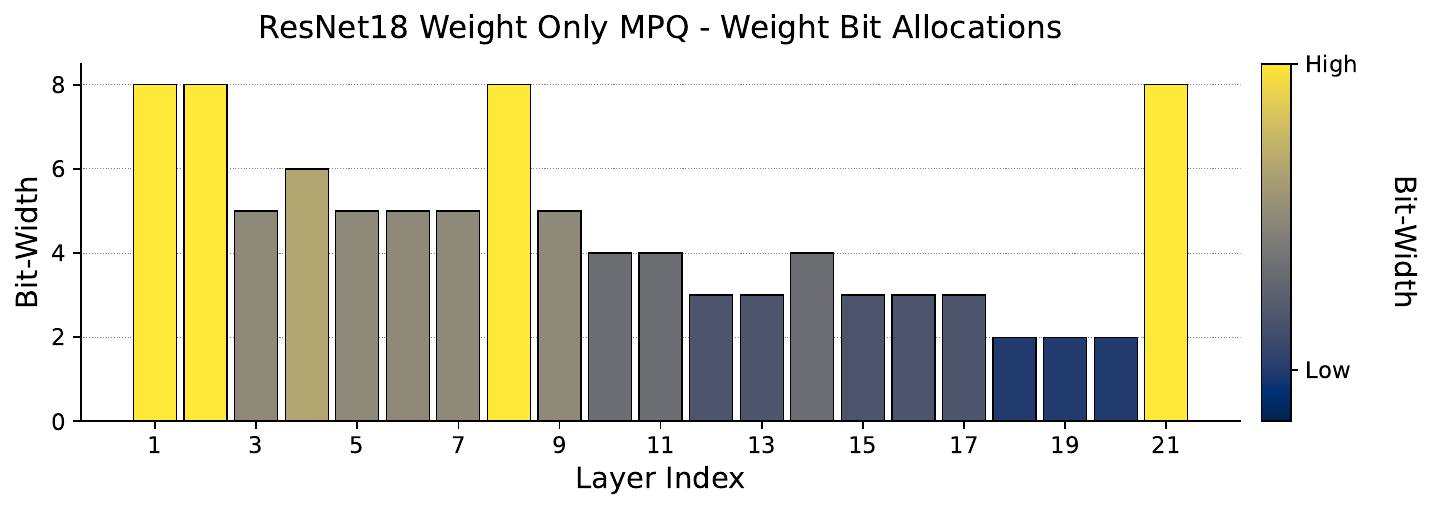}
    \caption{ResNet18 Weight Only Experiment Weight Bit Allocations}
    \label{fig:resnet18-wonly-bits}
\end{figure*}

\begin{figure*}
    \centering
    \begin{subfigure}{1\columnwidth}
            \includegraphics[width=1\linewidth]{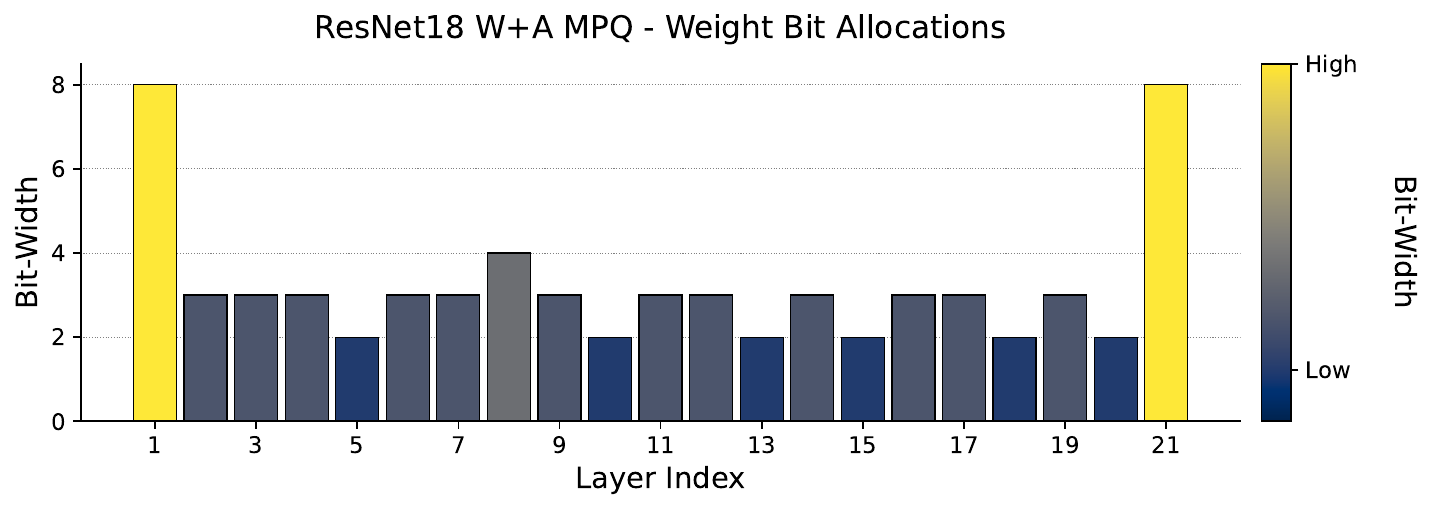}

    \end{subfigure}
    \begin{subfigure}{1\columnwidth}
        \includegraphics[width=1\linewidth]{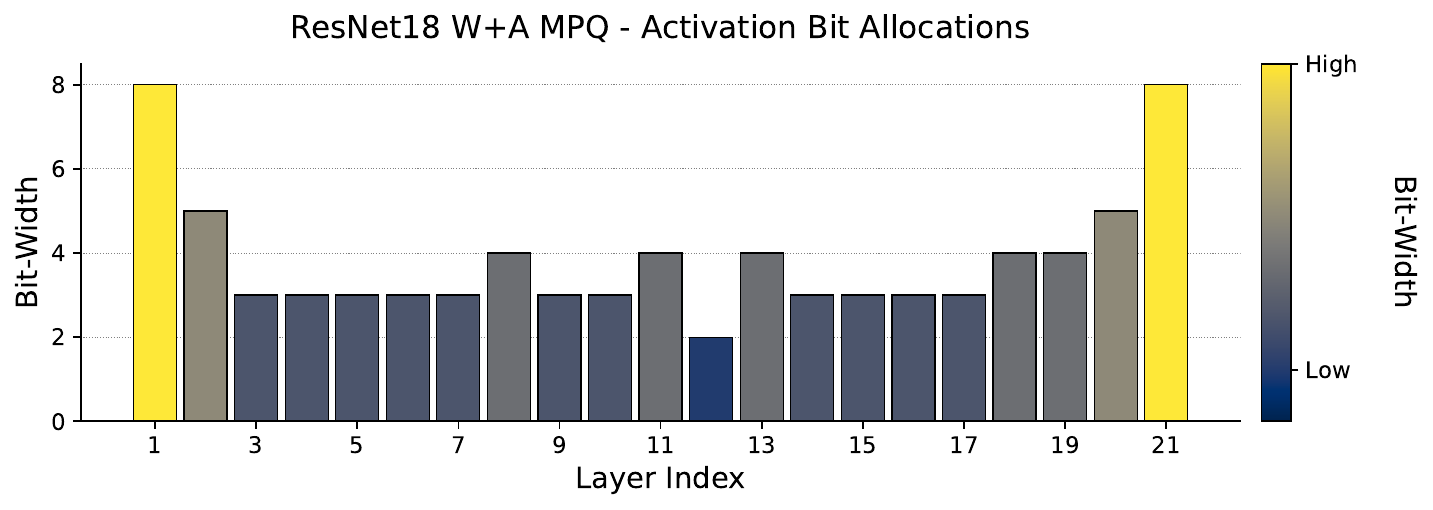}

    \end{subfigure}
\caption{ResNet18 Weight and Activation Bit Allocations. 'W+A' means weight and activation mixed precision quantization.}
\label{fig:resnet18_w+abits}
\end{figure*}

\begin{figure*}
    \centering
    \begin{subfigure}{1\columnwidth}
    \includegraphics[width=1\linewidth]{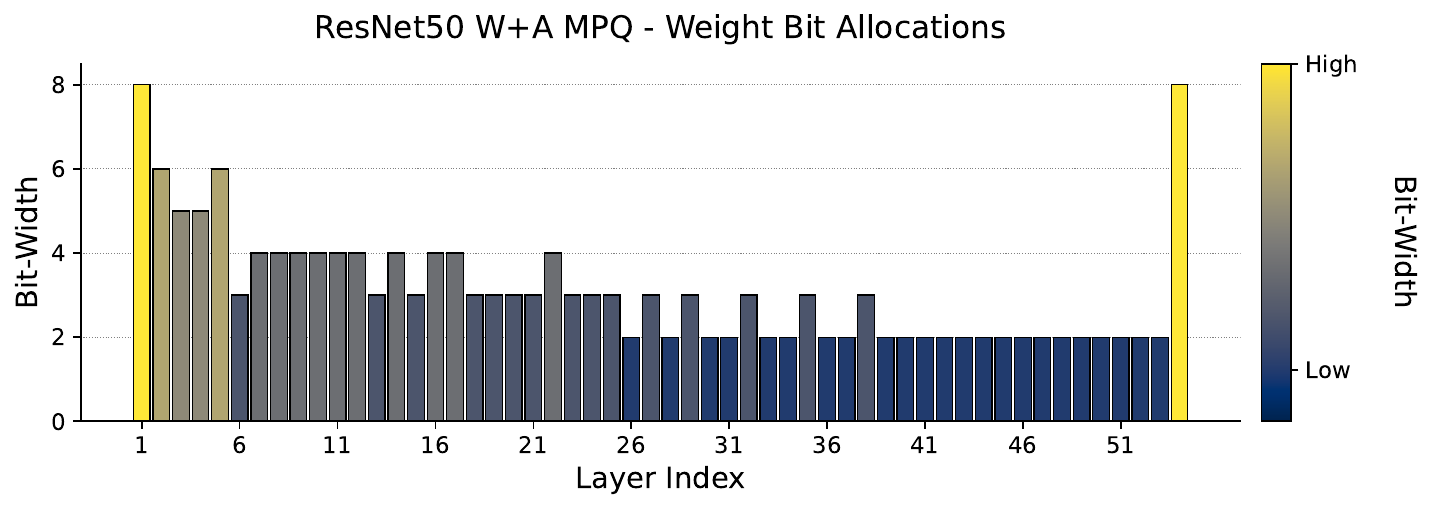}

    \end{subfigure}
    \begin{subfigure}{1\columnwidth}
    \includegraphics[width=1\linewidth]{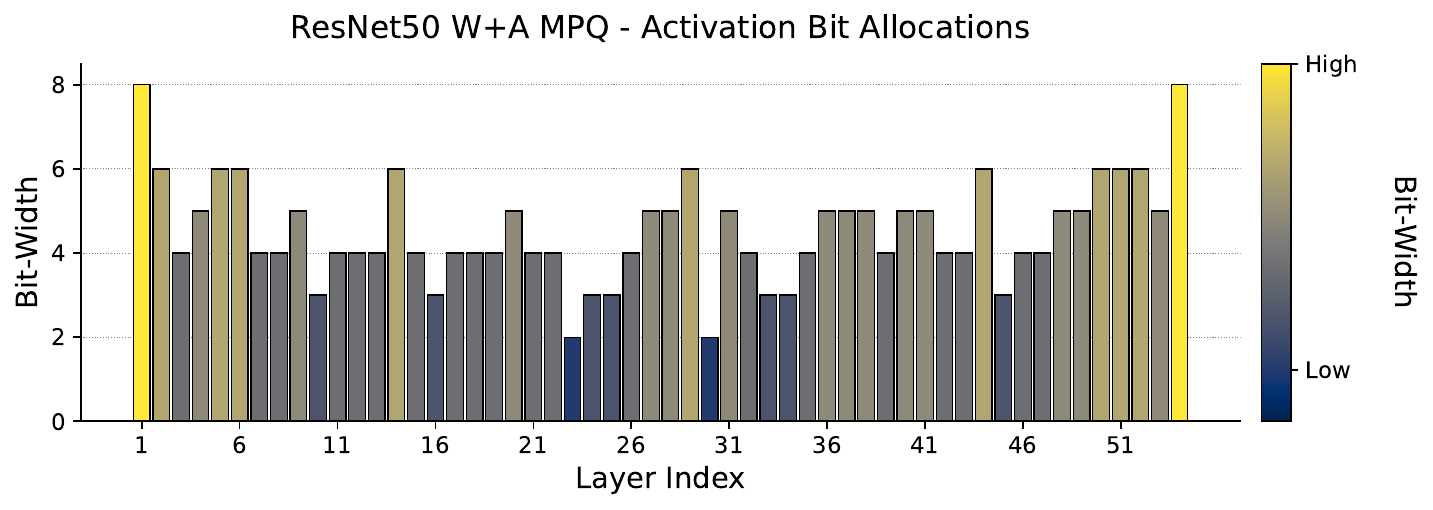}

    \end{subfigure}
\caption{ResNet50 Weight and Activation Bit Allocations. 'W+A' means weight and activation mixed precision quantization.}
\label{fig:resnet50_w+abits}
\end{figure*}

\begin{figure*}
    \centering
    \includegraphics[width=0.5\linewidth]{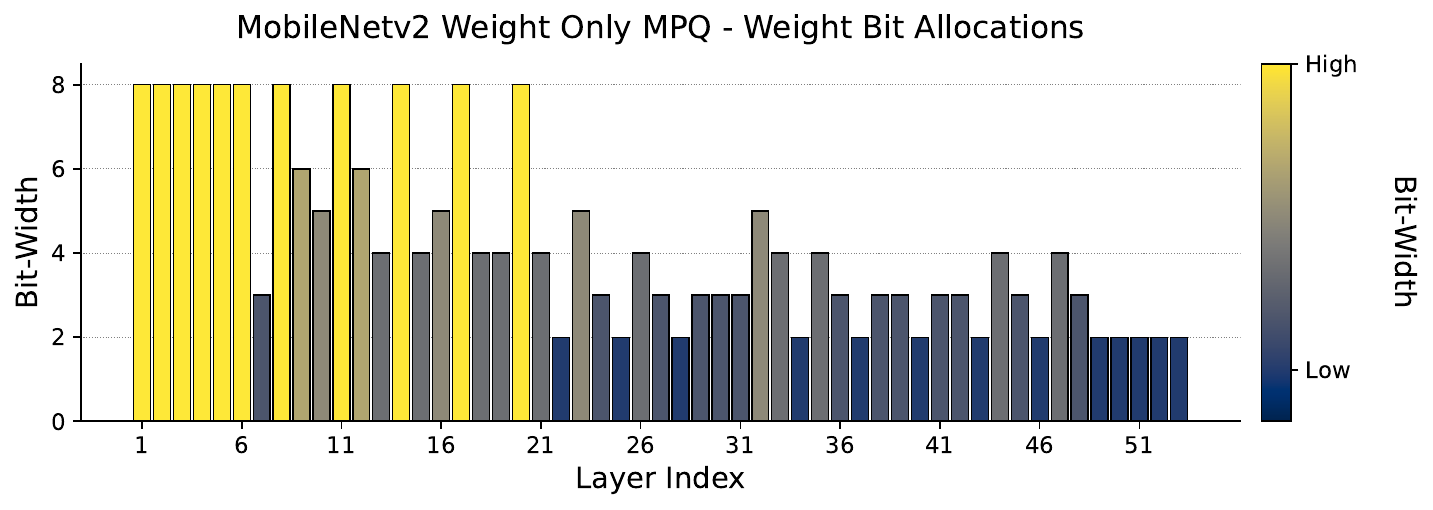}
    \caption{MobileNetv2 Weight Only Experiment Weight Bit Allocations}
    \label{fig:mobilenetv2_bits}
\end{figure*}

To provide full transparency for our experimental setup, we present a detailed breakdown of the bit-width allocations for the models under consideration, namely ResNet-18, ResNet-50, and MobileNetV2. These comprehensive allocations are visualized in Figures~\ref{fig:resnet18-wonly-bits}, \ref{fig:resnet18_w+abits}, \ref{fig:resnet50_w+abits}, and~\ref{fig:mobilenetv2_bits}. This observed distribution of precision is not random; rather, it aligns with a widely accepted heuristic in the field of mixed-precision quantization, i.e. that the initial layers of a network are highly sensitive to quantization noise. Consequently, they require higher bit-widths to preserve this critical information and prevent significant error propagation. In contrast, the final convolutional layers exhibit greater tolerance to the precision loss introduced by aggressive, low-bit quantization.

\clearpage
\section{Hyperparameter $a$}

The hyperparameter $\alpha$ serves to balance the relative importance of weight versus activation quantization sensitivity within our ILP formulation (Eq. 5). It is crucial to emphasize that this hyperparameter was utilized in only two specific experiments (on ResNet18 and ResNet50) where both weights and activations were quantized simultaneously. This was done to ensure a direct and fair comparison against our main competitor, which employs a similar balancing parameter. For all other Post-Training Quantization (PTQ) and weight-only Quantization-Aware Training (QAT) experiments presented in this work, this hyperparameter was not employed.

Despite its limited use, we provide a targeted ablation study to demonstrate that our method is not highly sensitive to the precise choice of $\alpha$. We performed this analysis on the ResNet50 model under a fixed model size constraint (12.2$\times$). For each value of $\alpha$ tested, we first determined the optimal bit-width configuration and then ran QAT for 10 epochs using a learning rate of 0.01 and a subset of 6,000 samples of ImageNet. This entire process was repeated three times for each $\alpha$ value.

The results of this study are summarized in Table~\ref{tab:alpha_ablation}. The data clearly indicates that variations in $\alpha$ within a reasonable range have a negligible impact on the final model performance. The mean Top-1 and Top-5 accuracies remain remarkably stable across the tested values, with minimal deviation. Furthermore, the low standard deviation for each set of runs confirms the consistency of the results. This study demonstrates that while $\alpha$ provides a mechanism for fine-tuning the balance between weight and activation sensitivity, the overall performance of InfoQ is robust and not dependent on meticulous tuning of this parameter.

\begin{table}[h!]
\centering
\label{tab:alpha_ablation}
\begin{tabular}{|c|c|c|}
\hline
\textbf{Value of $\boldsymbol{\alpha}$} & \textbf{Top-1 Acc. (\%)} & \textbf{Top-5 Acc. (\%)} \\
\hline
0.5 & $43.96 \pm 0.16$ & $70.05 \pm 0.22$ \\
1.0 & $44.06 \pm 0.17$ & $70.11 \pm 0.12$ \\
2.0 & $43.91 \pm 0.18$ & $69.96 \pm 0.21$ \\
3.0 & $44.04 \pm 0.27$ & $70.08 \pm 0.23$ \\
\hline
\end{tabular}
\caption{Ablation study on the hyperparameter $\alpha$ for ResNet50 after 10 epochs of QAT. The results show that both Top-1 and Top-5 accuracy are stable across different values of $\alpha$, indicating low sensitivity to this parameter.}
\label{tab:alpha_ablation}
\end{table}


\end{document}